\documentclass{article}

\usepackage[numbers,square,sort&compress]{natbib}

\usepackage[main, final]{neurips_2026}

\usepackage[utf8]{inputenc} % allow utf-8 input
\usepackage[T1]{fontenc}    % use 8-bit T1 fonts
\usepackage{url}            % simple URL typesetting
\usepackage{booktabs}       % professional-quality tables
\usepackage{amsfonts}       % blackboard math symbols
\usepackage{nicefrac}       % compact symbols for 1/2, etc.
\usepackage{microtype}      % microtypography

\usepackage[hidelinks]{hyperref}  % [hidelinks]
\usepackage[acronym,toc,shortcuts]{glossaries}
\glsdisablehyper
\usepackage{graphicx,wrapfig}
\usepackage{lipsum}
\usepackage{subcaption}
\usepackage{nameref}
\usepackage{siunitx}  % For aligning numbers by decimal point (S column type)
\usepackage{multirow}
\usepackage{caption}
\usepackage{comment}
\usepackage[table]{xcolor}
\usepackage{bm}
\usepackage{tikz}

\usepackage[frozencache=true,cachedir=minted-cache]{minted}
\usepackage{pifont}
\usepackage{amsmath}
\usepackage{amssymb}
\usepackage{mathtools}
\usepackage{amsthm}
\usepackage{makecell}
\usepackage[capitalize,noabbrev]{cleveref}

\newcommand{\ie}{\textit{, i.e., }}
\newcommand{\eg}{\textit{, e.g., }}

\newacronym{simba}{SimBa}{Simple Baseline}
\newacronym{simbamm}{SimBaMM}{Simple Baseline for Multimodal Learning}
\newacronym{lf}{LF}{Late Fusion}
\newacronym{ef}{EF}{Early Fusion}
\newacronym{moe}{MoE}{Mixture of Experts}
\newacronym{sota}{SoTA}{state-of-the-art}
\newacronym{flops}{FLOPS}{Floating Point Operations}
\newacronym{dit}{DiT}{Diffusion Transformer}
\newacronym{bce}{BCE}{Binary Cross Entropy}
\newacronym{vae}{VAE}{Variational Auto Encoder}
\newacronym{ukb}{UKB}{UK Biobank}
\newacronym{ecg}{ECG}{electrocardiogram}
\newacronym{auroc}{AUROC}{area under the receiver operating characteristic}
\newacronym{vit}{ViT}{Vision Transformer}
\newacronym{ehr}{EHR}{electronic health record}
\newacronym{ct}{CT}{computer-assisted tomography}
\newacronym{mlp}{MLP}{Multi-Layer Perceptron}
\newacronym{cnn}{CNN}{Convolutional Neural Network}
\newacronym{vlm}{VLM}{Vision-Language Model}
\newacronym{prs}{PRS}{polygenic risk score}
\newacronym{sd}{SD}{standard deviation}
\newacronym{rope}{ROPE}{Region of Practical Equivalence}

\title{Fusion or Confusion?\\Multimodal Complexity Is Not All You Need}
\author{
  \textbf{Tillmann Rheude\textsuperscript{1,3$^\dagger$}}, 
  \textbf{Roland Eils\textsuperscript{1,2,3$^\dagger$}}, 
  \textbf{Benjamin Wild\textsuperscript{1,4$^\dagger$}}\\
  \textsuperscript{1}Berlin Institute of Health, Charité - Universitätsmedizin Berlin,
  \textsuperscript{2}Intelligent Medicine Institute,\\Fudan University,
  \textsuperscript{3}Department of Mathematics and Computer Science, Freie Universität Berlin,
  \\\textsuperscript{4}Institute of Medical Informatics, Charité - Universitätsmedizin Berlin \\
  $^\dagger$\texttt{\{benjamin.wild, roland.eils, tillmann.rheude\}@bih-charite.de}
}

\begin{document}

\maketitle

\begin{abstract}
    % Topic 
    Multimodal learning has become a prominent research area, with the potential of substantial performance gains by combining information across modalities.
    % Motivation: Multimodal complexity is increased
    At the same time, model development has trended toward increasingly complex deep learning architectures, motivated by the assumption that multimodal-specific methods improve performance.
    % Contribution: Re-implementation of 19 methods, on 9 datasets, with up to 23 modalities 
    We challenge this assumption through a large-scale empirical study by reimplementing \num{19} high-impact \ac{lf} methods across nine diverse datasets with up to \num{23} modalities.
    % Detail / Nuance: to show that multimodal complexity is not needed given standardized weight initializations, hyperparameter tunings, cross-validations, not only for multimodal methods but also for unimodal baselines.
    Under standardized experimental conditions, including hyperparameter tuning, weight initialization, cross-validation, and statistical testing, increased multimodal complexity often yields \emph{confusion rather than effective fusion} of data modalities.
    Accordingly, complex multimodal architectures do not reliably outperform unimodal baselines and a \textbf{Sim}ple \textbf{Ba}seline for \textbf{M}ulti\textbf{m}odal Learning (\acs{simbamm}).
    % narrow impact: our case study shows that there are core methodological shortcomings in top-tier ML publications
    Through a focused case study, we further demonstrate concrete methodological shortcomings even in top-tier multimodal learning publications, underscoring the need for standardized evaluation practices.
    % broad impact: Away from the pursuit of architectural novelty
    In summary, we argue for a shift in focus for multimodal learning: away from the pursuit of architectural novelty and toward methodological rigor.\footnote{The code repository is uploaded to \href{https://github.com/TillmannRheude/simple_mml_baseline}{GitHub}.}
    %\footnote{The code repository will be available on GitHub and is attached anonymously for the review.} 
    
\end{abstract}

\section{Introduction}
\label{sec:introduction}

% Multimodal Learning 
Multimodal learning leverages information from multiple data modalities to improve predictive performance of deep learning architectures. This is especially important for domains with many modalities such as medical applications \cite{Acosta_Falcone_Rajpurkar_Topol_2022}. Therefore, multimodal learning can enrich unimodal representations and enable models to capture more complex patterns than any single modality alone.
% Multimodal Complexity
At the same time, the pursuit of \ac{sota} performance in multimodal learning has led to increasingly complex architectures. The field is engaged in an architectural arms race, driven by the implicit assumption that novel fusion mechanisms, gradient-based regularizations, and deeper, more sophisticated networks are the primary drivers of progress. This has resulted in a landscape in which models are often celebrated for their structural novelty, with performance gains on specific datasets presented as evidence of their superiority. 
% Statistics 101
However, such gains are difficult to interpret when evaluation protocols differ across papers and when key experimental factors\eg hyperparameter tuning, weight initialization, and missing-modality simulation, are not controlled or present at all.

In this paper, we challenge the prevailing narrative that architectural novelty is the primary driver of performance in multimodal learning. We argue that methodological rigor in model selection and evaluation can explain a substantial portion of reported improvements. Accordingly, performance gains which are attributed to complex architectural contributions may not reliably persist in fair comparisons. We hypothesize that multimodal learning beyond image-text pairs currently does not reliably benefit from increasingly complex architectural machinery on top of a simple \ac{lf} baseline. 
While multimodal learning can in principle improve downstream performance and should often contain more task-relevant information than any single modality alone, we argue that this potential is currently not captured by more complex multimodal methods.

We perform a large-scale empirical study evaluating \num{19} multimodal architectures across nine datasets with up to \num{488131} samples and \num{23} modalities. Our goal is to disentangle the contributions of architectural complexity from those of experimental practice. 
Given this large-scale effort of implementing, tuning and fairly comparing these multimodal methods, our findings reveal that increased architectural complexity does not yield consistent gains over a \ac{simbamm}. 
Rather than asking only in a benchmark-manner which multimodal architecture performs best, we ask under which evaluation conditions such conclusions are scientifically justified. 
Further, from a machine learning perspective, this finding is striking: In a literature where new multimodal architectures are repeatedly reported as superior to unimodal models, standard baselines, and earlier multimodal methods, these advantages often fail to persist under careful evaluation. If such claims are accepted without rigorous controls, the field risks optimizing for apparent novelty rather than for reliable scientific progress. Taken together, our key contributions are:

\begin{figure*}[t]
    \centering
    \includegraphics[width=1.0\textwidth]{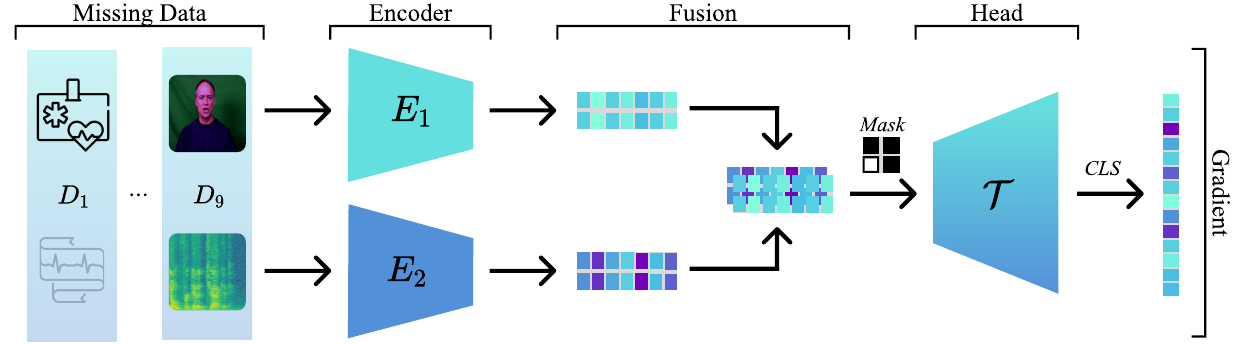}
    \caption{Taxonomy of multimodal learning methods exemplified by two modalities. An \ac{lf} baseline (\acs{simbamm}) processes a multimodal dataset $D$ with up to $M$ modalities through encoders $E_m$, combines them via fusion, and produces predictions through a classification head. 
    Categorized into five groups, we evaluate \num{19} \ac{sota} methods across nine datasets based on their architectural innovations on top of \acs{simbamm}: 
    \textbf{Missing Data} handling for incomplete modalities\eg a missing \acs{ecg}, \textbf{Encoder}-level modifications, \textbf{Fusion} mechanisms for combining representations, \textbf{Head}-based cross-modal architectures, and \textbf{Gradient}-based optimization techniques. Our work reveals that none of these complex extensions reliably outperforms \acs{simbamm}, suggesting that reported performance gains are often attributable to experimental confounders rather than methodological novelty.}
    \label{fig:late_fusion_groups}
\end{figure*}

\begin{itemize}
    \item \textbf{A large-scale analysis of multimodal architectures under a unified protocol}
    We reimplement and evaluate \num{19} methods across nine datasets with up to \num{23} modalities for controlled comparisons in low- and large-data regimes even beyond the original evaluation settings.
    
    \item \textbf{A \acf{simbamm}} 
    We show that recent multimodal methods are not able to outperform well-tuned unimodal baselines or a \acs{simbamm} (\Cref{fig:late_fusion_groups}) given fair comparisons and statistical correctness. 
    
    \item \textbf{A case study revealing how experimental confounders inflate performance}
    We highlight common multimodal evaluation weaknesses, such as information leakage during validation and inadequate hyperparameter tuning, by including a case study of a recent method. This shows how confounders can produce misleading impressions of progress, motivating a shift toward more careful and statistically grounded evaluation in machine learning.
\end{itemize}
\section{Related Work}
\label{sec:related_work}

\paragraph{Multimodal Learning}
Multimodal learning addresses the use of multiple modalities to perform prediction tasks. We focus on architectural contributions to multimodal learning rather than specific applications and we also exclude vision-language models since they pose a special case of multimodal learning. Architectures in multimodal deep learning can be divided into groups (\Cref{fig:late_fusion_groups}) spanning 
missing-data handling
\citep{Wang_Li_Cui_2023,Nezakati_Reza_Patil_Solh_Asif_2024,Wang_Chen_Ma_Avery_Hull_Carneiro_2023}, 
encoder  
\citep{Ghahremani_Wachinger_2023,Jiang_Huang_Yang_2025} and  
fusion mechanisms 
\citep{Nagrani_Yang_Arnab_Jansen_Schmid_Sun_2021,Liu_Shen_Lakshminarasimhan_Liang_Zadeh_Morency_2018,Cao_Xia_Ding_Zhang_Hu_2024,Li_Zhou_Yu_Song_Yang_2024,Wu_Shao_Wang_Sun_2025}, 
head models 
\citep{Tsai_Bai_Liang_Kolter_Morency_Salakhutdinov_2019}, and 
gradient regulation methods 
\citep{Peng_Wei_Deng_Wang_Hu_2022,Wei_Luo_Luo_2025a,Wei_Luo_Luo_2025b,Wei_Hu_2024,Wu_Jastrzebski_Cho_Geras_2022,Wang_Tran_Feiszli_2020}. 
There is also work on meta-learning \citep{Ma_Ren_Zhao_Tulyakov_Wu_Peng_2021} and similar extensions such as pre-training or distillation \citep{Li_Selvaraju_Gotmare_Joty_Xiong_Hoi_2021} to multimodal learning, which are excluded in our work.
Recent work conducted a large-scale study of scaling laws for native multimodal models \citep{Shukor_Fini_Costa_Cord_Susskind_El_Nouby_2025}. This line of work both supports and contextualizes our approach: their results motivate our use of an \ac{lf} transformer, while our study targets a different gap by investigating contemporary multimodal methods across diverse datasets with statistically rigorous evaluation. 
Furthermore, our work is situated within the context of multimodal benchmarking, with MultiBench \citep{Liang_Lyu_Fan_Wu_Cheng_Wu_Chen_Wu_Lee_Zhu_etal._2021} as the most closely related effort. However, we extend and differentiate our work from this in several ways. 
First, we evaluate generalizability on a curated set of large-scale, contemporary real-world datasets. Second, we assess cross-domain performance directly (rather than splitting in- vs. out-of-domain) to attribute gains to methodological contributions and test their transferability \citep{Lipton_Steinhardt_2019}. Third, we standardize key implementation details such as weight initializations.
This provides an immediate and actionable baseline and overview for the community, which is notably absent as the MultiBench leaderboard remains unreleased since its publication. Finally, we do not only benchmark methods but ask under which conditions supposed performance gains are scientifically justified.

\paragraph{Multimodal Learning with Missing Modalities}
Prior work on robustly handling missing modalities is broadly classified into two categories: strategy design (which includes architecture-focused designs and model combinations) and data processing (which involves representation learning and modality imputation) \citep{Wu_Wang_Chen_Carneiro_2024}. If missing modalities are not imputed directly at the data level, prior knowledge about missing modalities is used to engineer modules for handling missing modalities. This can\eg be done with latent space reconstructions \citep{Wang_Li_Cui_2023} or projections \citep{Nezakati_Reza_Patil_Solh_Asif_2024}, and distribution alignments \citep{Wang_Chen_Ma_Avery_Hull_Carneiro_2023}. On the other hand, Transformers \citep{Vaswani_Shazeer_Parmar_Uszkoreit_Jones_Gomez_Kaiser_Polosukhin_2017} are able to handle missing modalities since they can handle arbitrary sequence lengths. 
Even if prior work reports performance drops with missing modalities and finds that standard Transformers degrade in this setting \citep{Ma_Ren_Zhao_Testuggine_Peng_2022}, balanced training with missing modalities can be sufficient for downstream performance \citep{Memmel_Bachmann_Zamir_2023}. Yet, a large-scale, domain-independent study of such a \ac{simba} is still lacking.

\paragraph{Reproducibility of Machine Learning Methods} 
Machine learning suffers from a reproducibility crisis \citep{Kapoor_Narayanan_2023,Semmelrock_Ross_Hellauer_Kopeinik_Theiler_Haberl_Thalmann_Kowald_2025}. There is prior work focusing on reproducibility in healthcare \citep{McDermott_2025} and guidelines for reproducibility in life sciences \citep{Heil_Hoffman_Markowetz_Lee_Greene_Hicks_2021}. More generally, venues for high-impact machine learning publications are aware of this crisis\ie by offering solutions such as reproducibility challenges and reports \citep{Pineau_Vincent_Lamarre_Sinha_Lariviere_Beygelzimer_dAlche_Buc_Fox_Larochelle_2021}. 
Nevertheless, we consider not only reproducibility, but also unsound comparisons\eg missing hyperparameter optimizations similar to recent learning rate specific work \cite{Lee_Ko_Chen_Yeh_2026}.

\paragraph{\acfp{simba}}  
There is a variety of \acp{simba} for diverse tasks.
For example, for computer vision, there are \acp{simba} for 
video restoration \citep{Li_Shi_Zhang_Cheung_See_Wang_Qin_Li_2023}, 
tracking \citep{Lin_Fan_Zhang_Xu_Ling_2022},
image classification \citep{Chan_Jia_Gao_Lu_Zeng_Ma_2015,Liu_Mao_Wu_Feichtenhofer_Darrell_Xie_2022}, 
video text spotting \citep{He_Ye_Zhang_Liu_Du_Tao_2024},
hand mesh reconstruction \citep{Zhou_Zhou_Lv_Zou_Tang_Liang_2024},
graph generation \citep{Shi_Zhong_Xu_Li_Xu_2021},
video matting \citep{Li_Ohanyan_Goel_Navasardyan_Wei_Shi_2023},
spoken-to-sign-language translation \citep{Zuo_Wei_Chen_Mak_Yang_Tong_2024},
and
point cloud classification \citep{Goyal_Law_Liu_Newell_Deng_2021}. 
More general approaches tackle\eg 
active learning \citep{Kirsch_Farquhar_Atighehchian_Jesson_Branchaud_Charron_Gal_2023},
Bayesian uncertainty \citep{Maddox_Izmailov_Garipov_Vetrov_Wilson_2019},
transfer learning \citep{Chen_Wei_Sun_Wu_Lin_2022},
continual learning \citep{Buzzega_Boschini_Porrello_Abati_CALDERARA_2020,Prabhu_Torr_Dokania_2020,Press_Schneider_Kummerer_Bethge_2023,Zhang_Pfahringer_Frank_Bifet_Lim_Jia_2022},
the traveling salesman problem \citep{Xia_Yang_Liu_Liu_Song_Bian_2024}, 
time series \citep{Chen_Luong_Mukherjee_Singh_2025},
knowledge tracing \citep{Liu_Liu_Chen_Huang_Luo_2023}, and
instruction fine-tuning \citep{Zhao_Andriushchenko_Croce_Flammarion_2024}. 
In the context of multimodal learning, there is prior work related to 
multimodal image segmentation \citep{Bastico_Ryckelynck_Corte_Tillier_Decenciere_2023},
multimodal medical reasoning \citep{Huang_Wu_Liu_Tang_Zhou_2025},
multimodal embeddings \citep{Liang_Lim_Tsai_Salakhutdinov_Morency_2019},
multimodal representation learning \citep{Manolache_Tantaru_Niepert_2024},
multimodal \acp{vlm} \citep{Xu_Liu_He_Huang_Jiang_2024},
multimodal learning with \ac{moe} models \citep{Li_Chen_Han_2025}, and
multimodal domain generalization \citep{Dong_Nejjar_Sun_Chatzi_Fink_2023}.
Prior work hints that there is no benefit of multimodal extensions\eg fusion models \citep{Song_Chen_Jaume_Vaidya_Baras_Mahmood_2024}, more complex methods \citep{Caranzano_Pancotti_Rollo_Sartori_Lio_Fariselli_Sanavia_2025}, or pre-training in genomics \citep{Vishniakov_Viswanathan_Medvedev_Kanithi_Pimentel_Rajan_Khan_2024}. In summary, such an analysis for multimodal learning across all architectural components is missing.
\section{Method}
\label{sec:method}
Consider a dataset $D=\{( \{x_{i,m}\}_{m\in\mathcal{M}_i},\, y_i)\}_{i=1}^N$ consisting of $N$ samples.
Each sample $i$ is a (sub)set of $M$ possible modalities and a corresponding target label $y_i$.
The set of available modality indices is $\mathcal{M}_i \subseteq \{1,\dots,M\}$.
For an available modality $m\in\mathcal{M}_i$, the input is $x_{i,m}$.

\subsection{\Acf{simbamm}}
\label{sec:simbamm}
The \ac{simbamm} architecture processes each sample through a multi-stage \ac{lf} pipeline (\Cref{fig:late_fusion_groups}).

\paragraph{Encoders} 
For each available modality $m \in \mathcal{M}_i$, the input $x_{i,m}$ is passed through a modality-specific encoder $E_m$ to obtain an embedding
$e_{i,m} = E_m(x_{i,m})$.
Each $e_{i,m}$ is projected into a shared hidden space of dimension $d$ using a modality-specific linear layer $P_m$:
$\hat{e}_{i,m} = P_m(e_{i,m}) \in \mathbb{R}^{d}$.

\paragraph{Fusion}
We form an input sequence $S_i$ from the projected embeddings.
To balance performance and fine-grained modeling, we study two variants:
(i) \textbf{SimBaMM}, which uses a configurable number of tokens per modality (determined by the encoder output and a pooling/tokenization choice),
and (ii) \textbf{SimBaMM$^{\text{CLS}}$}, which represents each modality by a single token\eg a \texttt{[CLS]} embedding, yielding a fixed multimodal sequence length of $M$ with padding.

\paragraph{Head}
The head module is a Transformer $\mathcal{T}$ operating over $S_i$.
To handle missing modalities, we construct an attention mask $M_{\text{attn}}(\mathcal{M}_i)$ such that attention is only computed among tokens originating from available modalities. This enables a controlled comparison between mask-only handling and the additional mechanisms used in prior work.
The fused representation is
$H_i = \mathcal{T}(S_i, M_{\text{attn}}(\mathcal{M}_i))$.
Afterwards, a projection $g(\cdot)$ maps the fused representation $H_i$ to a task-specific output $\hat y_i= g(H_i)$.
We optimize a task loss $\mathcal{L}(y_i,\hat y_i)$\ie (binary) cross-entropy.

\subsection{Comparison Taxonomy, Method Selection \& Reimplementations}
\label{sec:method_selection_and_reimplementations}
We group multimodal methods by the type of innovation they introduce beyond \ac{lf}. \Cref{fig:late_fusion_groups} summarizes \ac{simbamm} and the comparison groups used throughout the experiments. 
We benchmark \ac{simbamm} against a representative set of recent and influential multimodal methods to test whether architectural and objective-level innovations provide consistent gains under a rigorous evaluation protocol.
Methods were selected to cover the comparison taxonomy in \Cref{fig:late_fusion_groups} based on:
(a) recency (primarily methods from the last few years),
(b) publication in top-tier peer-reviewed machine learning venues, and
(c) explicit architectural or objective-based modifications beyond a standard \ac{lf} baseline.
We prioritize methods with sufficiently detailed specifications and/or official reference implementations to enable faithful re-implementations. However, personal selection bias is difficult to eliminate entirely. We made every effort to choose representative methods without regard to expected outcomes.
%Full details of all \num{19} compared multimodal architectures and re-implementations are provided in \Cref{app:method_reimplementation_details}.
%All models are implemented in a shared codebase using PyTorch Lightning \citep{Falcon_PyTorch_Lightning_2019} to ensure consistent training, logging, and evaluation.
Encoders are dataset-specific but held fixed across all methods (\Cref{tab:dataset_summary}). Details in \Cref{app:method_reimplementation_details}. 
\begin{table*}[t]
    \centering
    \caption{
        Overview of datasets and the respective encoders used in our evaluation. Details in \Cref{app:dataset_details}. 
    }
    \label{tab:dataset_summary}
    \small
    \sisetup{group-separator={,}} % Optional: adds commas to large numbers
    \begin{tabular}{@{}l c c c c@{}}
    \toprule
    \textbf{Dataset} & 
    \textbf{Size} & 
    \textbf{Modalities} & 
    \textbf{Raw Input} & 
    \textbf{Encoders} \\
    \midrule
    Crema-D 
        & $7,442$ & 2 (Video, Audio) & \ding{51} & $2\times$ ResNet \\
    MIMIC HAIM
        & $45,050$ & 2 (Laboratory, X-ray) & \ding{51} & \acs{mlp}, \acs{vit} \\
    INSPECT
        & $22,449$ & 2 (\acs{ct}, \acs{ehr}) & \ding{55} & \acsp{mlp} \\
    MIMIC Symile
        & $10,345$ & 3 (Laboratory, X-ray, ECG) & \ding{51} & \acs{mlp}, \acs{vit}, Transformer \\
    CMU-MOSI
        & $2,199$ & 3 (Video, Audio, Language) & \ding{55} & $2\times$ Transformer,  BERT \\
    CMU-MOSEI 
        & $22,856$ & 3 (Video, Audio, Language) & \ding{55} & $2\times$ Transformer,  BERT \\
    CH-SIMS   
        & $2,281$ & 3 (Video, Audio, Language) & \ding{51} & ViViT, Wav2Vec2, BERT \\
    CH-SIMS 2 
        & $4,403$ & 3 (Video, Audio, Language) & \ding{51} & ViViT, Wav2Vec2, BERT \\
    \acs{ukb} 
        & $488,131$
        & \makecell{
            23 (\textit{e.g.}, Metabolomics, \acs{ehr}, \\ \acs{prs}, Proteomics, Biochemistry)
        } 
        & \ding{51} \ding{55} 
        & \acsp{mlp} \\

    %VGG 
    %    & $199,175$ & 2 & Video, Audio & \ding{55} & \acsp{mlp} \\
    %Kinetics 
    %    & $288,244$ & 2 & Video, Audio & \ding{55} & \acsp{mlp} \\
    \bottomrule
\end{tabular}
\end{table*}

\paragraph{Generalization to heterogeneous tasks}
Recent publications for multimodal methods typically assume a fixed number of modalities, specific modality types, and a single downstream task.
We adapt all methods to (i) an arbitrary number of modalities, (ii) heterogeneous modality types, and (iii) different tasks and loss functions while preserving the original method’s defining components.
If a method introduces auxiliary losses, we include them; if auxiliary-loss weighting is not specified, we introduce scalar weights and tune them under the same protocol as other hyperparameters.

\paragraph{Optimization and initialization}
To reduce the degrees of freedom and improve comparability, we adopt a common optimizer choice whenever compatible with the method. By default, we use the ScheduleFree \citep{Defazio_Yang_Khaled_Mishchenko_Mehta_Cutkosky_2024} variant of AdamW and SGD only when required by the method. Weight initializations are standardized across methods to ensure stable convergence across all methods. Details in \Cref{app:case_study_details}.

\subsection{Experimental Protocols}
\label{sec:experimental_protocols}
%\begin{table}[t]
\begin{wraptable}[9]{r}{0.4\textwidth}
    \centering
    \footnotesize
    \vspace{-0.47cm}
    \caption{
        Overview of evaluation protocols used by recent multimodal methods.
        Values are the percentage of surveyed multimodal papers using each protocol.
    }
    
    \label{tab:protocol_summary}
    \begin{tabular}{@{} l c @{}}
    \toprule
    \textbf{Evaluation Protocol} & \textbf{\% of Methods $\uparrow$} \\
    \midrule
    
    Hyperparameter Tuning & $0$ \\
    
    Statistical Significance & $0$ \\
    
    Cross-Validation & $11$ \\
    % RegBN
    % AUG
    
    Error Bars & $21$ \\
    % AUG
    % RegBN
    % PDF
    % BMML
    
    Unimodal Baselines & $82$ \\
    % IMDer
    % MMP
    % ShaSpec
    % RegBN
    % AUG
    % MBT
    % PDF
    % MulT
    % OGM
    % DGL
    % ARL
    % MMPareto
    % BMML
    % G-Blend
    
    \bottomrule
    \end{tabular}
%\end{table}
\end{wraptable}

We argue that multimodal methods can only be compared meaningfully under controlled experimental conditions. Therefore, we standardize the key components of training and evaluation across datasets and methods. The resulting protocols aim to reduce experimental confounding and make differences between methods scientifically interpretable. 
While most evaluated methods report unimodal baselines, other evaluation protocols are applied inconsistently (\Cref{tab:protocol_summary}). Moreover, unimodal baselines are not tuned similar to the proposed multimodal methods.

\paragraph{Missing-modality protocol}
We simulate missing modalities at the dataset-split level (rather than per batch) to ensure consistent and reproducible training conditions across epochs \citep{Rheude_Eils_Wild_2025}.
Concretely, we pre-generate a binary missingness mask for all samples in a split. 
This avoids pre-training or leakage on all available data modalities (\Cref{subsec:missing_modality_analysis}). 
Missing rates are defined per number of missing modalities: for a trimodal dataset, a rate of \num{15}\% means \num{15}\% of samples miss exactly one and another \num{15}\% miss exactly two modalities. 

\paragraph{Hyperparameter tuning protocol}
We first conduct a Bayesian hyperparameter search for \ac{simbamm} over a fixed set of parameters and an extensive search space to determine robust base configurations\eg encoder/head dimensions.
These base settings are then used as a starting point for tuning the remaining methods, focusing on method-specific components and optimizer parameters under a controlled tuning budget (\Cref{app:hyperparameter_search}). As a counterexample, we tune the full hyperparameter space of one method independently of \ac{simbamm} and find that this exhaustive per-method tuning does not materially change the results or conclusions (\Cref{app:case_study_details}).  

\paragraph{Bayesian statistical comparison} We employ Bayesian hierarchical analysis following \citet{Corani_Benavoli_Demsar_Mangili_Zaffalon_2017,Benavoli_Corani_Demsar_Zaffalon_2017} to compare methods across datasets (\Cref{app:statistical_comparison}). This allows us to account for correlations between cross-validation folds and estimate the posterior probability of three outcomes: (1) method $A$ outperforms $B$ by $>1\%$, (2) method $B$ outperforms $A$ by $>1\%$, or (3) they are practically equivalent (difference within a $\pm 1\%$ \ac{rope}).

\section{Experiments}
\label{sec:experiments}

\subsection{Multimodal Datasets}

We evaluate all methods on nine real-world multimodal datasets (\Cref{tab:dataset_summary}), spanning \num{2199} to \num{488131} samples and \num{2} to \num{23} modalities. This diverse collection enables a broad and controlled analysis of multimodal architectural complexity across substantially different modality regimes and dataset scales. We intentionally exclude synthetic or otherwise unrealistic datasets in order to ground our conclusions in practically relevant settings. The datasets used in this work are:
MIMIC Symile \citep{Saporta_Puli_Goldstein_Ranganath_2024}, MIMIC Haim \citep{Soenksen_Ma_Zeng_Boussioux_Villalobos_Carballo_Na_Wiberg_Li_Fuentes_Bertsimas_2022a,Soenksen_Ma_Zeng_Boussioux_Villalobos_Carballo_Na_Wiberg_Li_Fuentes_Bertsimas_2022b}, INSPECT \citep{Huang_Huo_Steinberg_Chiang_Lungren_Langlotz_Yeung_Shah_Fries_2023}, \ac{ukb} \citep{Sudlow_Gallacher_Allen_Beral_Burton_Danesh_Downey_Elliott_Green_Landray_2015}, MOSI \citep{Zadeh_Zellers_Pincus_Morency_2016}, MOSEI \citep{Zadeh_Liang_Poria_Cambria_Morency_2018}, CH-SIMS \citep{Yu_Xu_Meng_Zhu_Ma_Wu_Zou_Yang_2020}, CH-SIMS v2 \citep{Liu_Yuan_Mao_Liang_Yang_Qiu_Cheng_Li_Xu_Gao_2022}, and Crema-D \citep{Cao_Cooper_Keutmann_Gur_Nenkova_Verma_2014}.
%VGG-Sound \citep{Chen_Xie_Vedaldi_Zisserman_2020}, and Kinetics \citep{Kay_Carreira_Simonyan_Zhang_Hillier_Vijayanarasimhan_Viola_Green_Back_Natsev_et_al_2017}.
These datasets differ in both their input modalities and data representations, while remaining non-synthetic and representative of real-world tasks. For the \ac{ukb}, missing modalities are an inherent characteristic. Therefore, we implemented zero imputation for this dataset except for the missing data-based methods. Input modalities are z-score normalized for all datasets. Depending on the modality, we use a \ac{mlp}, \ac{vit} \citep{Dosovitskiy_Beyer_Kolesnikov_Weissenborn_Zhai_Unterthiner_Dehghani_Minderer_Heigold_Gelly_etal_2021}, Transformer \citep{Vaswani_Shazeer_Parmar_Uszkoreit_Jones_Gomez_Kaiser_Polosukhin_2017}, BERT \citep{Devlin_Chang_Lee_Toutanova_2019}, Wav2Vec2 \citep{Baevski_Zhou_Mohamed_Auli_2020}, ViViT \citep{Arnab_Dehghani_Heigold_Sun_Lucic_Schmid_2021}, or ResNet \citep{He_Zhang_Ren_Sun_2016} as encoders. We report mean and \ac{sd} using subject-wise five-fold cross-validation. This eliminates subject overlap between folds, addressing both a frequent methodological weakness in the field (see \Cref{subsec:case_study}) and the ability to generalize to new samples. Details in \Cref{app:dataset_details}.

\subsection{Downstream Task Analysis}
\label{subsec:downstream_task_analysis}
We provide a comprehensive empirical downstream comparison of \ac{simbamm} against the \num{19} re-implemented \ac{sota} methods, including unimodal performances in \Cref{fig:forest_plot} and \Cref{app:detailed_model_results}. 
The forest plot demonstrates that no method is adding significant downstream gains given a dataset-specific $\pm1\%$ \ac{rope}. 
Further, there are even outlier methods which do not generalize to datasets with many modalities ($M \gg 2$). This is exemplified with MulT and LMF which become unstable in this setting. MulT scales quadratically, requiring $M(M-1)$ cross-modal transformers ($506$ for $M{=}23$), which is memory-prohibitive. LMF's multiplicative fusion leads to vanishing gradients and activation collapse (\acs{auroc} $\approx$ random), suggesting tensor-product fusion is ill-suited for $M\gg2$.
For a statistical, dataset-wide summarizing comparison of all methods, we aim to answer two critical questions (\Cref{sec:experimental_protocols} and Details in \Cref{app:statistical_comparison}):
\begin{figure}[t]
    \centering
    \includegraphics[width=\linewidth]{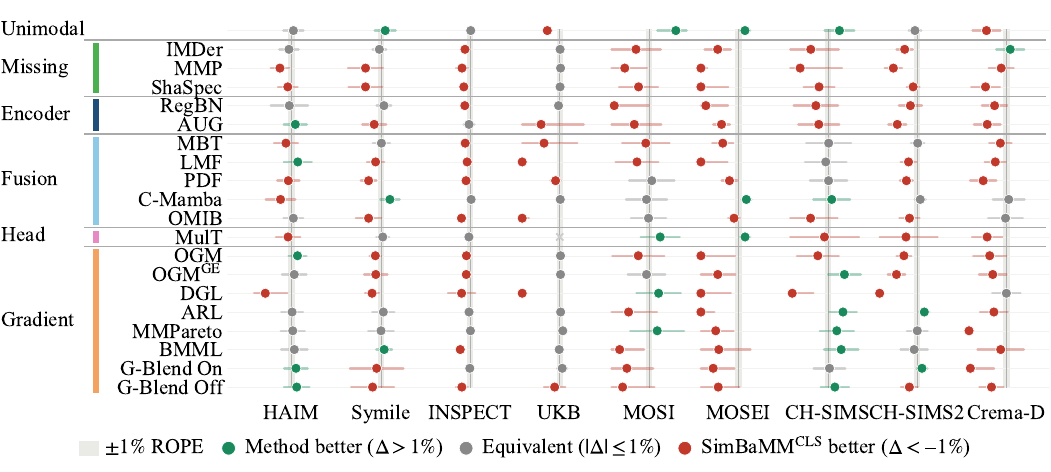}
    \caption{
        Aggregated forest plot comparing each method against \ac{simbamm}-CLS across all datasets. 
        Points show the mean performance difference relative to \ac{simbamm}-CLS, horizontal lines indicate uncertainty, and the shaded region marks the dataset-specific $\pm 1\%$ \ac{rope}, rather than the generalized \ac{rope} used in the other tables. 
        Unimodal represents the best-performing modality.
        No method shows a consistent aggregate advantage over \ac{simbamm}-CLS across datasets. Details in \Cref{app:detailed_model_results}.
    }
    \label{fig:forest_plot}
\end{figure}
%\begin{table}[t]
\begin{wraptable}[24]{r}{0.52\textwidth}
    \centering
    \footnotesize
    \setlength{\tabcolsep}{1.5pt}
    
    \caption{
        Comparison with \ac{simbamm}$^{\text{CLS}}$  (\emph{base} in this table). No  method achieves a high probability of
        meaningfully outperforming \ac{simbamm}$^{\text{CLS}}$. Details in \Cref{app:statistical_comparison}.
    }
    
    \label{tab:bayesian_simbamm_cls}
    \begin{tabular}{lccccc}
    \toprule
    \textbf{Method} & \makecell{\textbf{P(base}\\\textbf{$>$other)}} & \makecell{\textbf{P($\approx$)}} & \makecell{\textbf{P(other}\\\textbf{$>$base)}} & \makecell{\textbf{$\mathbb{E}[\delta_0]$}} & \makecell{\textbf{95\% CI}} \\
    \midrule
    Best Unimodal 
        & 0.53 & 0.06 & 0.41 & 0.002 & [-0.031,0.031] \\
    C-Mamba \citep{Li_Zhou_Yu_Song_Yang_2024}
        & 0.03 & 0.67 & 0.30 & -0.006 & [-0.021,0.006] \\
    MMPareto \citep{Wei_Hu_2024}
        & 0.13 & 0.67 & 0.19 & -0.002 & [-0.019,0.022] \\
    ARL \citep{Wei_Luo_Luo_2025b}
        & 0.30 & 0.47 & 0.23 & 0.002 & [-0.019,0.028] \\
    GBlend-On \citep{Wang_Tran_Feiszli_2020}
        & 0.36 & 0.48 & 0.16 & 0.005 & [-0.018,0.045] \\
    BMML \citep{Wu_Jastrzebski_Cho_Geras_2022}
        & 0.75 & 0.05 & 0.19 & 0.013 & [-0.021,0.051] \\
    OGM$^{\text{GE}}$ \citep{Peng_Wei_Deng_Wang_Hu_2022}
        & 0.77 & 0.16 & 0.06 & 0.015 & [-0.009,0.042] \\
    MulT \citep{Tsai_Bai_Liang_Kolter_Morency_Salakhutdinov_2019}
        & 0.33 & 0.41 & 0.26 & 0.002 & [-0.023,0.031] \\
    \ac{simbamm} 
        & 0.52 & 0.23 & 0.25 & 0.007 & [-0.038,0.056] \\
    MBT \citep{Nagrani_Yang_Arnab_Jansen_Schmid_Sun_2021}
        & 0.68 & 0.29 & 0.03 & 0.012 & [-0.004,0.028] \\
    GBlend-Off \citep{Wang_Tran_Feiszli_2020}
        & 0.90 & 0.06 & 0.04 & 0.021 & [-0.005,0.048] \\
    LMF \citep{Liu_Shen_Lakshminarasimhan_Liang_Zadeh_Morency_2018}
        & 0.89 & 0.01 & 0.10 & 0.049 & [-0.028,0.142] \\
    OMIB \citep{Wu_Shao_Wang_Sun_2025}
        & 0.95 & 0.04 & 0.01 & 0.022 & [0.003,0.043] \\
    OGM \citep{Peng_Wei_Deng_Wang_Hu_2022}
        & 0.85 & 0.12 & 0.03 & 0.017 & [-0.003,0.040] \\
    PDF \citep{Cao_Xia_Ding_Zhang_Hu_2024}
        & 0.94 & 0.05 & 0.01 & 0.020 & [0.005,0.037] \\
    AUG \citep{Jiang_Huang_Yang_2025}
        & 0.97 & 0.01 & 0.02 & 0.032 & [-0.000,0.066] \\
    RegBN \citep{Ghahremani_Wachinger_2023}
        & 0.75 & 0.22 & 0.03 & 0.014 & [-0.004,0.039] \\
    DGL \citep{Wei_Luo_Luo_2025a}
        & 0.98 & 0.00 & 0.02 & 0.078 & [0.006,0.158] \\
    \bottomrule
    \end{tabular}
%\end{table}
\end{wraptable}

\textbf{Does multimodal complexity yield performance gains?} Our analysis suggests limited evidence for practically meaningful performance gains from multimodal complexity. No complex method achieves a high probability of outperforming \ac{simbamm} (\Cref{tab:bayesian_simbamm_cls}). The single highest probability of a complex method winning is $\approx 30\%$ (C-Mamba). Conversely, for methods like AUG, DGL, and OMIB, \ac{simbamm}$^{\text{CLS}}$ is almost certainly superior ($P(\text{Base}>\text{Method}) \ge 95\%$). For the strongest competitors (C-Mamba, MMPareto), the data supports practical equivalence ($P_{\text{ROPE}} \approx 70\%$), indicating limited evidence for a consistent $\geq 1\%$ improvement over the baseline.

\textbf{Do multimodal methods improve upon unimodal baselines?} Given our empirical results, multimodal methods do not reliably improve upon unimodal baselines (\Cref{tab:bayesian_unimodal}). 
This is consistent with prior observations that the best unimodal model can at times outperform the joint multimodal model \cite{Wang_Tran_Feiszli_2020,Cadene_Dancette_Ben_younes_Cord_Parikh_2019,Gu_Fu_Liu_Valanarasu_Codella_Tan_Liu_Jin_Zhang_Wang_etal_2025}. However, in contrast to work that explicitly proposes mechanisms to mitigate such optimization or fusion difficulties \cite{Wang_Tran_Feiszli_2020}, we do not propose a dedicated solution here, but instead assess empirically whether the reported multimodal gains hold under a controlled and comparable evaluation setting.
While C-Mamba ($\approx 72\%$) and ARL ($\approx 67\%$) likely outperform the best unimodal baseline, many complex architectures\eg DGL, OMIB, LMF, and PDF are likely \textit{worse} than simply using the best single modality ($P(\text{Unimodal}>\text{Method}) > 80\%$). \ac{simbamm}$^{\text{CLS}}$ ($\approx 53\%$) rivals the top complex methods in its ability to outperform unimodal baselines, suggesting that architectural complexity does not reliably yield meaningful performance improvements for the multimodal datasets in our study.

\subsection{Efficiency Analysis}
\begin{figure}[t]
    \resizebox{\linewidth}{!}{
        \input{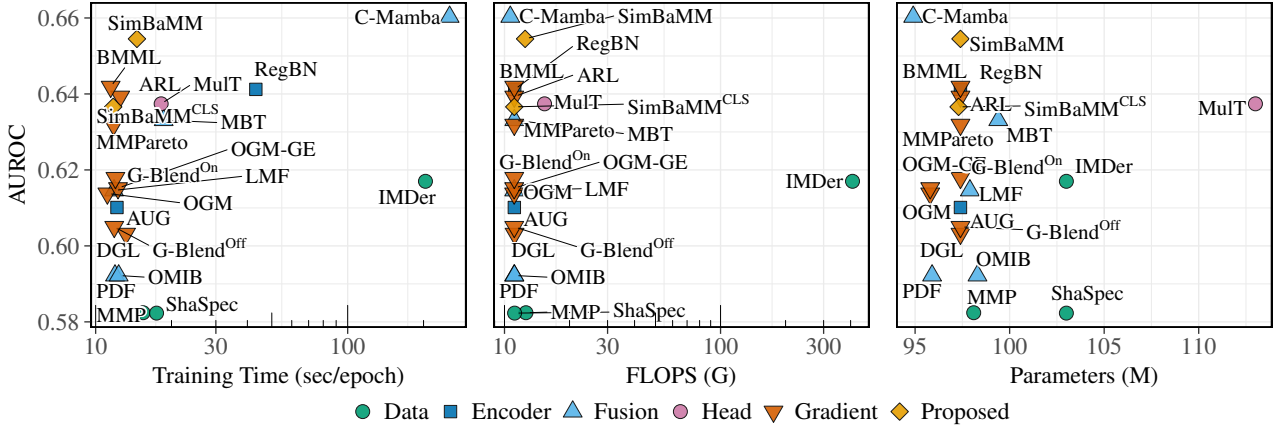}
    }
    \caption{Efficiency analysis for training time (left), FLOPS (middle), and parameters (right) on the MIMIC Symile dataset. Regardless of the metric, \ac{simbamm} and \ac{simbamm}$^{\text{CLS}}$ perform consistently while others\eg Coupled Mamba and IMDer are metric-dependent.}
    \label{fig:efficiency_analysis_detailed}
\end{figure}

We analyze the selected methods in terms of \ac{flops}, number of parameters, and runtime per epoch to demonstrate that many complex models incur high computational costs without yielding improvements in downstream performance (\Cref{fig:efficiency_analysis_detailed}). For instance, IMDer \cite{Wang_Li_Cui_2023} requires \num{203}s of runtime and \num{408}~G\ac{flops} because of the diffusion steps, yet only achieves an \acs{auroc} of \num{0.6170}.
Similarly, MulT, which has the largest number of parameters (\num{113}M), results in a lower \acs{auroc} of \num{0.6374} compared to more parameter-efficient models.
While C-Mamba reaches the highest \acs{auroc} (\num{0.6603}), it comes at the cost of a prohibitive runtime of \num{253}s.
In contrast, \ac{simbamm} demonstrates a balance of efficiency and effectiveness.
It achieves a high \acs{auroc} of \num{0.6545} while requiring only \num{14.6}s of runtime, \num{12.5}~G\ac{flops}, and \num{97.4}M parameters.
This is \num{17}$\times$ faster than C-Mamba for a minimal drop in \acs{auroc}.
These results underscore that \ac{simbamm} offers strong performance without the substantial computational overhead associated with other methods, positioning it as a practical and effective solution.
A parameter-matched analysis was not performed since it is non-trivial to determine how to equitably allocate a fixed parameter budget\eg whether to up- or downscale the encoders, the fusion module, or the classifier head. More details in \Cref{app:detailed_efficiency_analysis}.

\begin{wrapfigure}[15]{r}{0.52\textwidth}
    \centering
    \vspace{-15pt}
    \includegraphics[width=\linewidth]{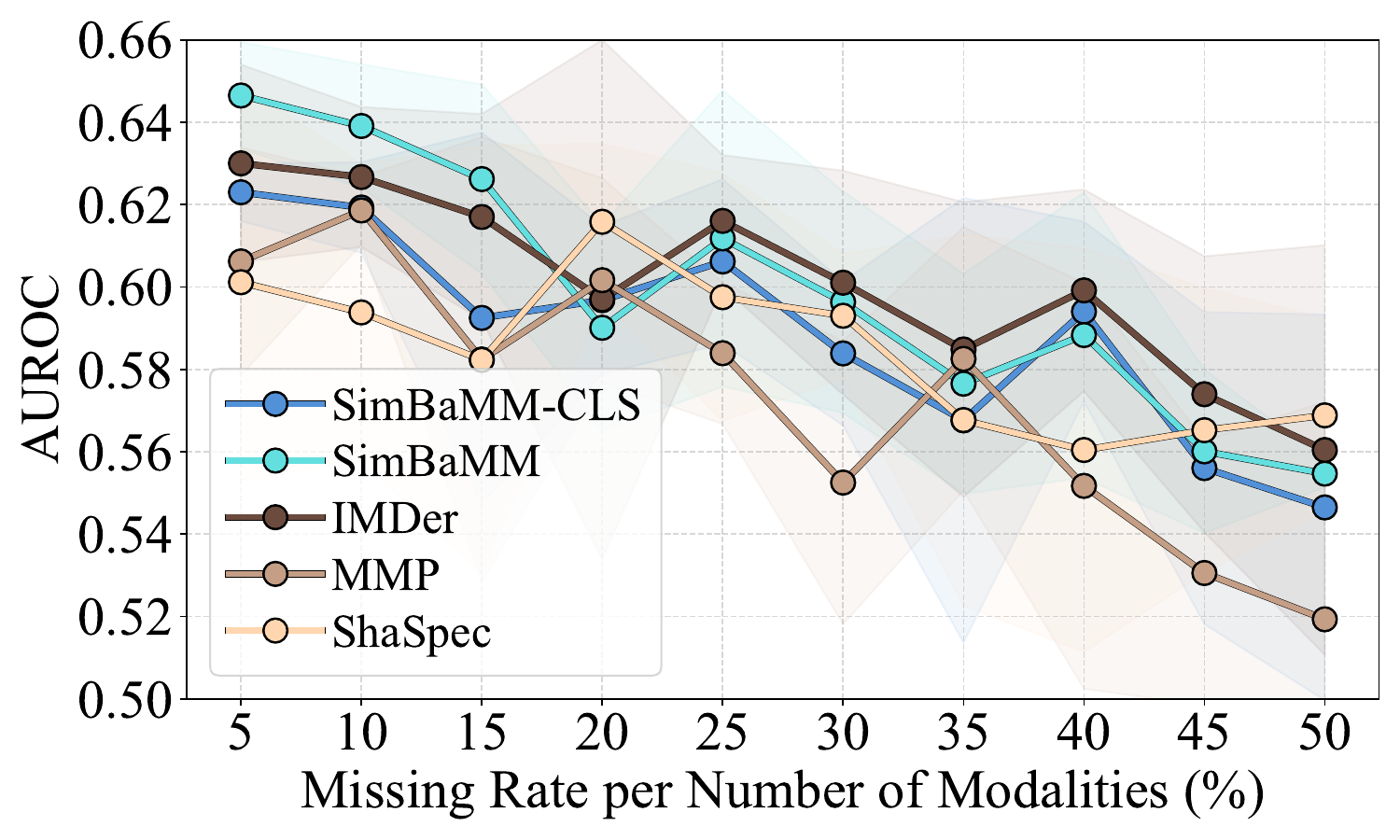}
    \caption{
        Missing modality analysis for IMDer \cite{Wang_Li_Cui_2023}, MMP \cite{Nezakati_Reza_Patil_Solh_Asif_2024}, ShaSpec \cite{Wang_Chen_Ma_Avery_Hull_Carneiro_2023}, and SimBaMM. Methods perform similarly across missing rates.
    }
    \label{fig:missing_modality_analysis}
\end{wrapfigure}
\subsection{Missing Modality Analysis}
\label{subsec:missing_modality_analysis}
Analyzing missing modalities is important because robustness under incomplete input is one of the motivations for multimodal learning. In practice, however, stronger performance of sophisticated methods under missingness cannot be taken for granted: In \Cref{fig:missing_modality_analysis}, we compare performance across different missing rates on MIMIC Symile and observe that the competing methods do not reliably outperform \ac{simbamm}. A plausible explanation is that some approaches are trained under assumptions that are not fully aligned with realistic deployment. For example, the public IMDer \cite{Wang_Li_Cui_2023} implementation instructs users to initialize missing-modality experiments from weights pretrained on complete multimodal data\ie first using the full modality set and then fine-tuning on the same dataset with missing modalities \cite{Wang_Li_Cui}, which may yield a less realistic evaluation protocol for missingness scenarios.

\subsection{Case Study}
\label{subsec:case_study}
\begin{wrapfigure}[17]{r}{0.52\textwidth}
    \centering
    \vspace{-13pt}
    \includegraphics[width=\linewidth]{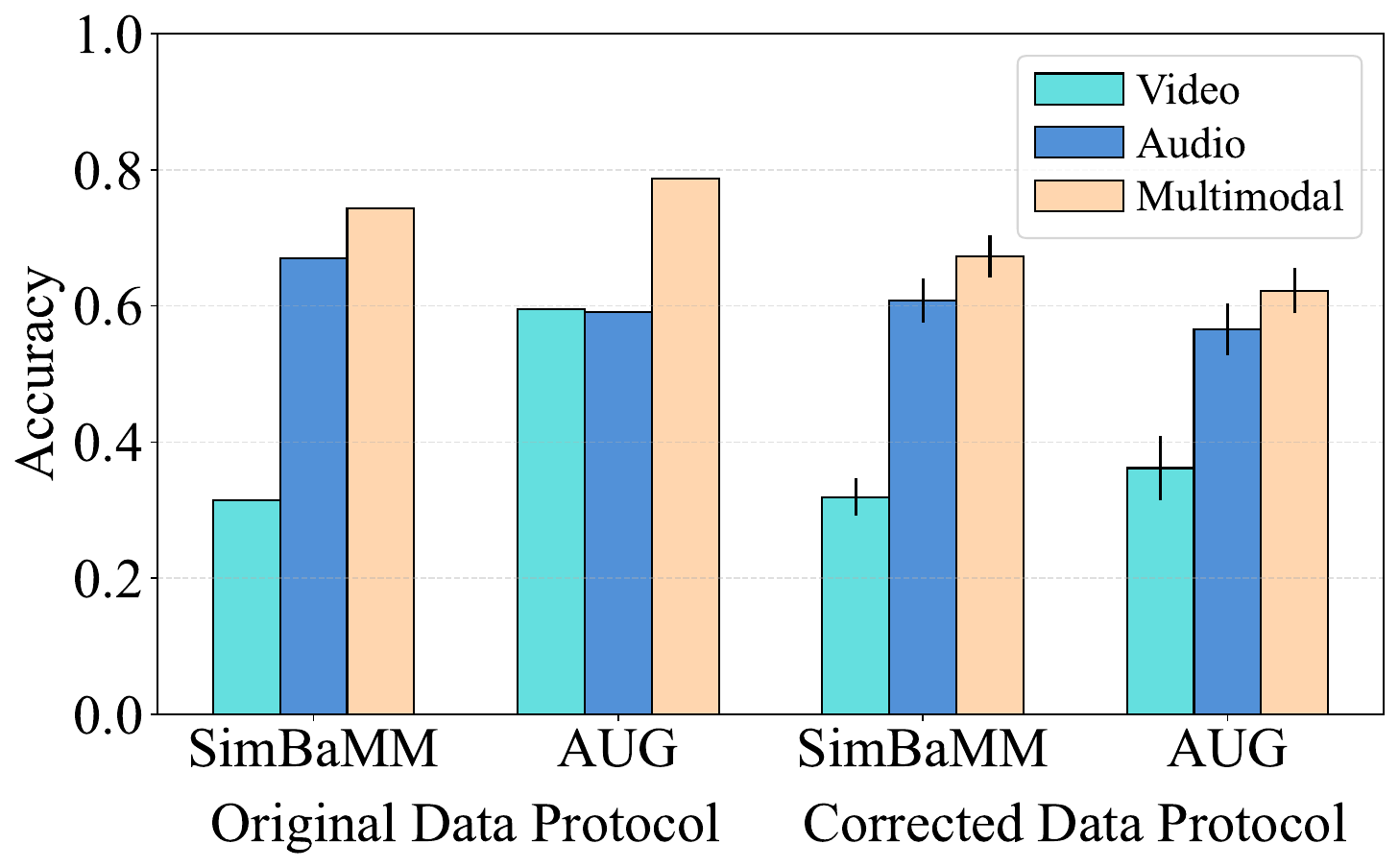}
    \caption{Case study on the CREMA-D dataset. Under a corrected data protocol with subject-independent splits and consistent hyperparameter tuning, the relative performance of methods changes.}
    \label{fig:case_study}
\end{wrapfigure}
We highlight methodological challenges that can compromise the validity of conclusions to illustrate how common evaluation practices can affect reported results.
The Crema-D dataset is widely used in multimodal learning, yet its official repository does not provide standardized cross-validation splits. It states only that subject-independence should be ensured when generating them, leaving this critical decision to individual researchers \cite{Wang_Cao_Cooper_Keutmann_Gur_Nenkova_Verma}.
We examine AUG\footnote{Published at NeurIPS 2025 with an oral presentation.} \citep{Jiang_Huang_Yang_2025} as a representative and recent example of how these issues can affect reported results. Based on the original publication, AUG appears to outperform a broad range of multimodal learning methods with an innovative boosting algorithm, supported by ablation studies, bar charts, sensitivity analyses, and t-SNE \citep{Maaten_Hinton_2008} plots. 
However, except for tabular results, these validations are reported exclusively on Crema-D. AUG follows the splits from OGM, which uses only training and test sets, resulting in model selection on the test set. Furthermore, the splits are not subject-independent despite the dataset repository's explicit guidance, resulting in subject leakage. 
Additionally, different optimizers are used across datasets, complicating fair comparisons. 
When we introduce a standardized dataset split protocol, generalize the implementation of AUG to $M$ modalities, and tune hyperparameters, we observe that AUG continues to boost unimodal video performance. However, it does not improve multimodal performance relative to simpler baselines. Consequently, the relative ordering shifts from AUG outperforming \ac{simbamm} to \ac{simbamm} matching or exceeding AUG (\Cref{fig:case_study}).
While acknowledging the theoretical contributions of AUG, this demonstrates how common evaluation practices can influence conclusions. Details in \Cref{app:case_study_details}.

\section{Discussion}
\label{sec:discussion}

\subsection{Limitations}
% Scope of the work 
\paragraph{Scope of the work}
Our conclusions should be interpreted within the scope of the present work. Although we evaluate $19$ multimodal methods across nine datasets, our study does not cover all multimodal learning paradigms such as pretraining. Therefore, \ac{ef} methods, large-scale pre-training-based approaches, contrastive objectives, and some domain-specific architectures fall outside the scope of this work. We do not claim that architectural complexity is never useful in multimodal learning, but only that we do not observe reliable practical gains over a strong \ac{simba} in the settings studied.
While the evaluated datasets span diverse modality combinations and up to $23$ modalities, they do not justify universal conclusions about all multimodal learning problems.
% Task settings and Biases
\paragraph{Task settings and biases}
Our work necessarily involves a trade-off between comparability and method-specific optimization. Some methods were originally proposed for narrower task settings or modality combinations and may therefore be disadvantaged. Similarly, our use of a standardized tuning protocol improves fairness and comparability, but may not be optimal for every architectural family. Starting from infrastructure choices that perform well for \ac{simbamm} may introduce a tuning bias, even though all methods are subsequently tuned with substantial compute budgets. 
We do not claim that each method performs identically under its original protocol. This would reintroduce multiple confounders. Our conclusions apply to the controlled setting studied here.
% Evaluation Protocols 
\paragraph{Evaluation protocols}
Our conclusions depend on concrete evaluation choices, including preprocessing, evaluation protocols, and the use of specific \acp{rope} in the Bayesian analysis. While we provide sensitivity analyses, alternative choices could affect absolute performance and rankings.
% Mechanistic insights
\paragraph{Mechanistic insights}
Our paper is primarily empirical and diagnostic. It establishes that carefully tuned \acp{simba} are difficult to outperform under controlled evaluation, but it provides only limited mechanistic insight into why some complex methods fail, why certain methods such as Coupled Mamba show positive trends, or which specific design principles should guide future multimodal architectures. We view these questions as important and promising directions for future work.

\subsection{Main Drivers of Multimodal Learning}
Given our empirical results that complex \ac{sota} multimodal models do not yield reliable performance gains, a critical question arises: If methodological novelty is not the answer, what are the true drivers of multimodal performance?
We argue that most problems framed as multimodal are, in fact, general machine learning challenges. Consider the core motivations behind the methods in our study: non-alignment, differential learning dynamics across modalities, dominant modality suppression, missing modality imputation, and distribution heterogeneity. These challenges arise equally in unimodal learning when combining heterogeneous feature groups, or when learning from features with varying predictive strength. The distinction between \emph{modalities} and \emph{feature groups within a modality} is largely semantic. If these problems do not require specific solutions in unimodal contexts, it is unclear why they would require such solutions in multimodal contexts.
This suggests that the performance of any multimodal architecture is anchored to the quality of its unimodal representations. For example, in unimodal learning, the paradigm shift from \acp{cnn} \citep{LeCun_Bengio_Hinton_2015} to \acp{vit} \citep{Dosovitskiy_Beyer_Kolesnikov_Weissenborn_Zhai_Unterthiner_Dehghani_Minderer_Heigold_Gelly_etal_2021} improved visual feature extraction. A multimodal model using a \ac{vit} backbone benefits from this, even before any fusion is applied. Similarly, innovations in attention\eg local, global, or windowed attention, are developed in unimodal contexts (\textit{e.g.}, the Global Context \ac{vit} \citep{Hatamizadeh_Yin_Heinrich_Kautz_Molchanov_2023}) before possibly being adapted to multimodal architectures. The gains from these components are substantial, yet they are not multimodal innovations. The true performance drivers may be the unimodal parts, while the specific choice of a more complex multimodal method may offer only marginal, if any, benefit.

\subsection{Reliability Checklist for Multimodal Learning}
\label{subsec:reliability_checklist}
We argue for a shift towards reliable results instead of added complexity. 
For comparable and trustworthy future work, we recommend the following pragmatic best practices: 

\textbf{(1) Standardize experimental conditions}: Use the same optimizers, weight initialization, data splits, and training protocols across all methods. \textbf{(2) Use identical base architectures}: When claiming benefit from a specific contribution, use an identical base architecture for both the proposed method and baselines. \textbf{(3) Include baselines}: Include simple and unimodal baselines such as \acs{simbamm} to verify that observed gains stem from the proposed contribution rather than confounding factors. \textbf{(4) Tune all methods rigorously}: Perform hyperparameter tuning for both the proposed method and baselines. \textbf{(5) Use proper cross-validation and reporting}: Employ cross-validation with subject-independent splits and report mean $\pm$ \ac{sd} rather than single-run results. \textbf{(6) Evaluate across multiple diverse datasets}: Assess generalizability by testing methods across multiple diverse and preferably large datasets.

%To make multimodal learning results comparable and trustworthy, we recommend the following minimal checklist. 
%All methods should be implemented in a single framework such as Lightning with shared data loaders, optimizers, schedulers, and logging. 
%For training and evaluation, fixed train/validation/test splits in combination with cross-validation ensuring no leakage in between splits should be used. 
%Unimodal baselines and a simple fusion baselines such as \ac{simbamm} should be included. For any added mechanism beyond this baseline, a cross-validated ablation study should be provided to verify benefit. 
%It should be at least a reasonably large grid or random search over key hyperparameters performed, selecting models based on validation performance only.
%Weight initialization should be regarded w.r.t. all layers of all architectures.
%Test performance should be reported as mean~$\pm$~standard deviation (or confidence intervals) over seeds or folds rather than single numbers.

\section{Conclusion \& Future Work}
\label{sec:conclusion_and_future_work}
We challenged the pursuit of architectural novelty in multimodal learning with a large-scale empirical study reimplementing \num{19} \ac{sota} models across nine diverse datasets with up to \num{23} modalities. Our analysis reveals that complex methods were at best practically equivalent to our proposed \ac{simbamm}, with none showing strong evidence of meaningful performance gains. Our findings motivate a shift in emphasis: from architectural novelty alone to rigorous evaluation that verifies generalizable improvements. In many settings we study, careful tuning and strong baselines explain most of the observed gains. We demonstrate that performance is more robustly driven by rigorous hyperparameter tuning, sound methodological evaluation, and strong unimodal encoders, rather than by multimodal architectural novelty. 
Future work could include a similar analysis of other model families, such as \ac{ef} and \ac{moe} architectures, as well as methods employing other training strategies like pre-training, contrastive learning, or meta-learning. Another objective could be to develop a unified, all-inclusive model based on \ac{simbamm}.
%  and to release the entire experimental setup as a public framework and leaderboard to track reproducible progress in the field

\paragraph{Acknowledgement} We would like to thank Georg v. Arnim for the support with preprocessing the INSPECT dataset. Further, we would like to thank Stefan Hegselmann, Tom Burgert, Leon Sixt and Jake Graving for their valuable feedback and discussions on this paper. 
The authors acknowledge the Scientific Computing of the IT Division at the Charité - Universitätsmedizin Berlin for providing computational resources that have contributed to the research results reported in this paper. 
This research has been conducted using the UK Biobank Resource under application number 49966. B.W. and R.E. acknowledge support by the Collaborative Research Center (SFB 1470) funded by the German Research Council (DFG). 
The authors gratefully acknowledge the computing time made available to them on the high-performance computer "Lise" at the NHR Center NHR@ZIB. This center is jointly supported by the Federal Ministry of Education and Research and the state governments participating in the NHR (\url{www.nhr-verein.de/unsere-partner}). % Electronic copies of these publications will be sent by e-mail to hpc-projects@nhr.zib.de.

\bibliography{references}
\bibliographystyle{plainnat}

\clearpage
\newpage 

\appendix

\section{Broader Impact and Ethics}
This paper presents work whose goal is to advance the field of machine learning. There are many potential societal consequences of our work, none of which we feel must be specifically highlighted here.

\section{Method Reimplementation Details}
\label{app:method_reimplementation_details}

For every method, weight initialization is applied throughout the entire architecture, including encoders and method-specific extensions (Kaiming initialization for linear and convolutional layers with biases set to zero, constant values for the weight (\num{1.0}) and bias (\num{0.0}) of LayerNorm layers, and Gaussian initialization for additional parameters). Model-related blocks\eg for Transformers or ResNets, are not re-initialized specifically. For Transformer blocks, sinusoidal positional encodings are used. Unless a method architecturally requires full token sequences\eg cross-attention between modality token sequences, we propagate a single modality token\eg a \texttt{[CLS]} representation, from each encoder for efficiency and to keep bandwidth comparable across methods.
%To improve reproducibility and ensure faithful re-implementations, we contacted the first and last author of all benchmarked methods by sharing our preprint and code repository, inviting them to review the corresponding re-implementation details and report any discrepancies.
%\input{sections/appendix/method_reimplementations}

\section{Compute Environment}
\label{app:compute_environment}
\newacronym{hpc}{HPC}{High-Performance Cluster}
Our experiments are conducted on a \ac{hpc} with the following environment:
\begin{itemize}
    \item 21 Dell PowerEdge R7525 compute nodes, each with 64 AMD Epyc cores (Rome), 512GB RAM and 1 NVIDIA A100 40G GPU
    \item 2 Dell PowerEdge XE8545 compute nodes, each with 128 AMD Epyc cores (Milan), 512GB RAM, 4 NVIDIA A100 40G GPUs (NVLink-connected)
\end{itemize}
All efficiency-related experiments are done on the first\ie single-GPU compute node.

\section{Detailed Model Results}
\label{app:detailed_model_results}
In addition to \Cref{fig:forest_plot}, \Cref{tab:results_emotion_incl_cremad,tab:results_healthcare} present the detailed model results across the datasets. For datasets without a sequence dimension, \ac{simbamm} reduces to \ac{simbamm}$^{\text{CLS}}$, resulting in n/a for the other \ac{simbamm} variant. 
\begin{table*}[ht]
    \centering
    \small
    \caption{
        Performance in terms of AUROC. Details including full unimodal results in \Cref{app:dataset_details}. No method is highlighted because performance differences fall within the reported mean $\pm$ \acp{sd}, summarized in \Cref{tab:bayesian_simbamm_cls}.
    }
    \label{tab:results_healthcare}
    
    \begin{tabular}{@{}l c c c c@{}}  
        \toprule
        
        \textbf{Method} & 
        {\textbf{HAIM} $\uparrow$} & 
        {\textbf{Symile} $\uparrow$} & 
        {\textbf{INSPECT} $\uparrow$} & 
        {\textbf{\acs{ukb}} $\uparrow$} \\
        \midrule

        \arrayrulecolor{black!40}

        %\multicolumn{5}{l}{\hspace{-6pt}\textit{Unimodal}} \\
        Best Unimodal 
            & $0.7042 \pm .020$
            & $0.6453 \pm .030$
            & $0.6568 \pm .003$
            & $0.7560 \pm .010$ \\
        \midrule 
            
        \multicolumn{5}{l}{\hspace{-6pt}\textit{Missing Data-Based}} \\
        IMDer \citep{Wang_Li_Cui_2023} 
            & ${0.6900 \pm .018}$ & $0.6266 \pm .016$ & ${0.6385 \pm .007}$ & $0.7969 \pm .003$ \\
        MMP \citep{Nezakati_Reza_Patil_Solh_Asif_2024}
            & $0.6611 \pm .016$ & $0.5824 \pm .054$ & $0.6289 \pm .018$ & $0.7984 \pm .004$ \\
        ShaSpec \citep{Wang_Chen_Ma_Avery_Hull_Carneiro_2023} 
            & $0.6863 \pm .019$ & $0.5823 \pm .052$ & $0.6348 \pm .006$ & $0.7968 \pm .007$ \\
        % & SMIL \citep{Ma_Ren_Zhao_Tulyakov_Wu_Peng_2021} & --- & --- & --- & --- & --- \\
        %& \color{red} EBR  \color{black} \citep{Chaudhuri_Dutta_Bui_Georgescu_2025} 
        %    & --- & $0.5621 \pm x$ & --- & --- \\
        %& \color{red} SimMLM \color{black} \citep{Li_Chen_Han_2025} 
        %    & --- & --- & --- & --- & --- \\
        
        \cmidrule(lr){2-5}
        \ac{simbamm}$^{\text{CLS}}$
            & $0.6558 \pm .031$ & $0.5925 \pm .045$ & ${0.6378 \pm .008}$ & ${0.7957 \pm .008}$ \\
        \ac{simbamm} 
            & ${0.6600 \pm .024}$ & ${0.6262 \pm .023}$ & n/a & n/a \\

        %\addlinespace
        \midrule
        %\addlinespace

        \multicolumn{5}{l}{\hspace{-6pt}\textit{Encoder-Based}} \\
        RegBN \citep{Ghahremani_Wachinger_2023} 
            & $0.6912 \pm .054$ & $0.6412 \pm .018$ & $0.6379 \pm .002$ & $0.7919 \pm .001$ \\
        AUG \citep{Jiang_Huang_Yang_2025}
            & $0.7103 \pm .027$ & $0.6101 \pm .035$ & $0.6513 \pm .016$ & $0.7361 \pm .136$ \\
        \midrule 

        %\multirow{1}{*}{\textit{Latents}} 
        %& --- & --- & --- & --- & --- & --- & --- & --- & --- \\
        %\midrule 

        \multicolumn{5}{l}{\hspace{-6pt}\textit{Fusion-Based}} \\
        %& ALBEF \citep{Li_Selvaraju_Gotmare_Joty_Xiong_Hoi_2021} 
        %    & --- & $0.6190 \pm x$ & --- & --- \\
        MBT \citep{Nagrani_Yang_Arnab_Jansen_Schmid_Sun_2021}
            & $0.6804 \pm .029$ & $0.6331 \pm .023$ & $0.6391 \pm .003$ & $0.7456 \pm .106$ \\
        %MCR \citep{Kontras_Strypsteen_Chatzichristos_Liang_Blaschko_Vos_2025} 
        %    & $0.7078 \pm .034$ & $0.6170 \pm .048$ & $0.6030 \pm .017$ & --- \\
        %& AV-MC \citep{Liu_Yuan_Mao_Liang_Yang_Qiu_Cheng_Li_Xu_Gao_2022} 
        %    & --- & --- & --- & --- \\
        LMF \citep{Liu_Shen_Lakshminarasimhan_Liang_Zadeh_Morency_2018} 
            & ${0.7181 \pm .037}$ & $0.6146 \pm .023$ & $0.6454 \pm .003$ & $0.4999 \pm .000$ \\
        PDF \citep{Cao_Xia_Ding_Zhang_Hu_2024} 
            & $0.6880 \pm .025$ & $0.5922 \pm .020$ & $0.6419 \pm .003$ & $0.7825 \pm .014$ \\
        Coupled Mamba \citep{Li_Zhou_Yu_Song_Yang_2024} 
            & $0.6631 \pm .040$ & $0.6603 \pm .027$ & $0.6580 \pm .004$ & $0.7974 \pm .003$ \\
        OMIB \citep{Wu_Shao_Wang_Sun_2025} 
            & $0.7040 \pm .019$ & $0.5922 \pm .038$ & $0.6272 \pm .008$ & $0.6262 \pm .070$ \\
        
        \midrule 

        \multicolumn{5}{l}{\hspace{-6pt}\textit{Head-Based}} \\
        MulT \citep{Tsai_Bai_Liang_Kolter_Morency_Salakhutdinov_2019}
            & $0.6872 \pm .030$ & $0.6374 \pm .014$ & $0.6510 \pm .004$ & OOM \\
        \midrule 

        \multicolumn{5}{l}{\hspace{-6pt}\textit{Gradient-Based}} \\
        OGM \citep{Peng_Wei_Deng_Wang_Hu_2022} 
            & $0.7172 \pm .014$ & $0.6138 \pm .011$ & $0.6431 \pm .005$ & $0.7982 \pm .013$ \\
        $\text{OGM}^{\text{GE}}$ \citep{Peng_Wei_Deng_Wang_Hu_2022} 
            & $0.7069 \pm .030$ & $0.6152 \pm .034$ & $0.6426 \pm .002$ & $0.7982 \pm .008$ \\
        DGL \citep{Wei_Luo_Luo_2025a} 
            & $0.6143 \pm .066$ & $0.6032 \pm .017$ & $0.6276 \pm .043$ & $0.5803 \pm .053$ \\
        ARL \citep{Wei_Luo_Luo_2025b} 
            & $0.7000 \pm .024$ & $0.6393 \pm .032$ & $0.6514 \pm .002$ & $0.7986 \pm .003$ \\
        MMPareto \citep{Wei_Hu_2024}
            & $0.7021 \pm .029$ & $0.6319 \pm .038$ & $0.6559 \pm .005$ & ${0.8054 \pm .009}$ \\
        BMML \citep{Wu_Jastrzebski_Cho_Geras_2022}  
            & $0.7072 \pm .034$ & $0.6419 \pm .020$ & $0.6233 \pm .003$ & $0.7940 \pm .003$ \\
        G-Blend Online \citep{Wang_Tran_Feiszli_2020} 
            & $0.7123 \pm .028$ & $0.6179 \pm .084$ & $0.6538 \pm .005$ & $0.8036 \pm .014$ \\
        G-Blend Offline \citep{Wang_Tran_Feiszli_2020} 
            & $0.7148 \pm .032$ & $0.6050 \pm .067$ & $0.6284 \pm .018$ & $0.7795 \pm .033$  \\
        %& \color{red} PMR \color{black} \citep{Fan_Xu_Wang_Wang_Guo_2023}     
        %    & --- & $0.6071 \pm x$ & --- & --- \\
        %& \color{red} EBR \color{black} \citep{Chaudhuri_Dutta_Bui_Georgescu_2025} 
        %    & --- & $0.6014 \pm x$ & --- & --- \\
        \midrule 

        \ac{simbamm}$^{\text{CLS}}$
            & ${0.6985 \pm .025}$ & $0.6318 \pm .015$  & ${0.6556 \pm .006}$ & ${0.7957 \pm .008}$ \\
        \ac{simbamm}
            & $0.6871 \pm .029$ & $0.6429 \pm .036$  & n/a & n/a \\

        \arrayrulecolor{black}
        
        \bottomrule
    \end{tabular}
    
\end{table*}

\begin{table*}[t]
    \centering
    \small
    \caption{
        Performance in terms of $\text{Acc}_{7}$ (MOSI/MOSEI), $\text{Acc}_{5}$ (CH-SIMS/CH-SIMS2) and  $\text{Acc}_{6}$ (Crema-D). No method is highlighted because performance differences fall within the reported mean $\pm$ \acp{sd}, summarized in \Cref{tab:bayesian_simbamm_cls}.
    }
    \label{tab:results_emotion_incl_cremad}
    
    \begin{tabular}{@{}l c c c c c@{}}  
        \toprule
        
        \textbf{Method} & 
        {\textbf{MOSI} $\uparrow$} & 
        {\textbf{MOSEI} $\uparrow$} & 
        {\textbf{CH-SIMS} $\uparrow$} & 
        {\textbf{CH-SIMS 2} $\uparrow$} &
        {\textbf{Crema-D} $\uparrow$} \\
        \midrule

        \arrayrulecolor{black!40}

        \multicolumn{6}{l}{\hspace{-6pt}\textit{Unimodal}} \\
        Language 
            & $0.4084 \pm .024$ 
            & $0.5144 \pm .006$ 
            & $0.5458 \pm .022$ 
            & $0.4280 \pm .014$
            & n/a \\
        Vision 
            & $0.2556 \pm .038$ 
            & $0.4285 \pm .007$ 
            & $0.3593 \pm .014$  
            & $0.2497 \pm .010$
            & $0.3192 \pm .027$ \\
        Audio 
            & $0.2401 \pm .011$ 
            & $0.4197 \pm .005$ 
            & $0.4298 \pm .021$ 
            & $0.2686 \pm .031$
            & $0.6081 \pm .032$ \\
        \midrule 

        \multicolumn{6}{l}{\hspace{-6pt}\textit{Missing Data-Based}} \\
        IMDer \citep{Wang_Li_Cui_2023} 
            & $0.2810 \pm .058$ & ${0.4273 \pm .039}$ & $0.4541 \pm .079$ & $0.3951 \pm .022$ & $0.6840 \pm .032$ \\
        MMP \citep{Nezakati_Reza_Patil_Solh_Asif_2024}
            & $0.2451 \pm .046$ & $0.2760 \pm .117$ & $0.4201 \pm .127$ & $0.3589 \pm .024$ & $0.6552 \pm .025$ \\
        ShaSpec \citep{Wang_Chen_Ma_Avery_Hull_Carneiro_2023}
            & $0.28921 \pm .031$ & $0.3489 \pm .111$ & ${0.4806 \pm .026}$ & ${0.4222 \pm .015}$ & $0.6051 \pm .034$ \\

        \cmidrule(lr){2-6}
        \ac{simbamm}$^{\text{CLS}}$
            & $0.2683 \pm .057$ & $0.3769 \pm .083$ & $0.4716 \pm .062$ & ${0.4059 \pm .017}$ & $0.6380 \pm .039$ \\
        \ac{simbamm} 
            & ${0.3028 \pm .060}$ & ${0.4169 \pm .076}$ & ${0.5122 \pm .047}$ & $0.4090 \pm .012$ & n/a \\

        %\addlinespace
        \midrule 
        %\addlinespace

        \multicolumn{6}{l}{\hspace{-6pt}\textit{Encoder-Based}} \\
        RegBN \citep{Ghahremani_Wachinger_2023} 
            & $0.2115 \pm .096$ & $0.3894 \pm .069$ & $0.4704 \pm .057$ & $0.4135 \pm .034$ & $0.6340 \pm .026$ \\
        AUG \citep{Jiang_Huang_Yang_2025}
            & $0.2756 \pm .067$ & $0.4395 \pm .022$ & $0.4797 \pm .050$ & $0.3715 \pm .025$ & $0.6105 \pm .029$ \\
        \midrule 

        \multicolumn{6}{l}{\hspace{-6pt}\textit{Fusion-Based}} \\
        MBT \citep{Nagrani_Yang_Arnab_Jansen_Schmid_Sun_2021}
            & $0.3124 \pm .053$ & $0.4434 \pm .030$ & $0.5113 \pm .058$ & $0.4367 \pm .016$ & $0.6525 \pm .018$ \\
        %MCR \citep{Kontras_Strypsteen_Chatzichristos_Liang_Blaschko_Vos_2025}
        %    & $0.2401 \pm .091$ & $0.1999 \pm .085$ & $0.5524 \pm .038$ & $0.4413 \pm .013$ \\
        %& AV-MC \citep{Liu_Yuan_Mao_Liang_Yang_Qiu_Cheng_Li_Xu_Gao_2022} 
        %    & --- & --- & --- & --- \\
        LMF \citep{Liu_Shen_Lakshminarasimhan_Liang_Zadeh_Morency_2018}
            & $0.2842 \pm .042$ & $0.3704 \pm .087$ & $0.5020 \pm .051$ & $0.4074 \pm .021$ & $0.6363 \pm .013$ \\
        PDF \citep{Cao_Xia_Ding_Zhang_Hu_2024}
            & $0.3315 \pm .048$ & $0.4649 \pm .020$ & $0.5109 \pm .042$ & $0.4006 \pm .015$ & $0.5976 \pm .028$ \\
        C-Mamba \citep{Li_Zhou_Yu_Song_Yang_2024}
            & $0.3147 \pm .037$ & ${0.5188 \pm .005}$ & $0.5209 \pm .041$ & $0.4448 \pm .010$ & $0.6792 \pm .040$ \\
        OMIB \citep{Wu_Shao_Wang_Sun_2025}
            & $0.3211 \pm .015$ & $0.4795 \pm .011$ & $0.4535 \pm .075$ & $0.4105 \pm .029$ & $0.6690 \pm .046$ \\
        
        \midrule

        \multicolumn{6}{l}{\hspace{-6pt}\textit{Head-Based}} \\
        MulT \citep{Tsai_Bai_Liang_Kolter_Morency_Salakhutdinov_2019}
            & ${0.3584 \pm .031}$ & $0.5147 \pm .006$ & $0.4979 \pm .102$ & $0.3993 \pm .100$ & $0.6102 \pm .037$ \\
        \midrule 

        \multicolumn{6}{l}{\hspace{-6pt}\textit{Gradient-Based}} \\
        OGM \citep{Peng_Wei_Deng_Wang_Hu_2022}
            & $0.2883 \pm .063$ & $0.3060 \pm .176$ & $0.4766 \pm .052$ & $0.3927 \pm .019$ & $0.6185 \pm .043$ \\
        $\text{OGM}^{\text{GE}}$ \citep{Peng_Wei_Deng_Wang_Hu_2022}
            & $0.3147 \pm .026$ & $0.4274 \pm .054$ & ${0.5617 \pm .031}$ & $0.3688 \pm .024$ & $0.6281 \pm .031$ \\
        DGL \citep{Wei_Luo_Luo_2025a}
            & $0.3533 \pm .046$ & $0.3253 \pm .143$ & $0.3948 \pm .053$ & $0.2471 \pm .039$ & $0.6717 \pm .033$ \\
        ARL \citep{Wei_Luo_Luo_2025b}
            & $0.2574 \pm .073$ & $0.2838 \pm .132$ & $0.5570 \pm .013$ & ${0.4579 \pm .008}$ & $0.6318 \pm .032$ \\
        MMPareto \citep{Wei_Hu_2024}
            & $0.3488 \pm .067$ & $0.4209 \pm .055$ & $0.5373 \pm .035$ & $0.4354 \pm .030$ & $0.4398 \pm .046$ \\
        BMML \citep{Wu_Jastrzebski_Cho_Geras_2022}
            & $0.2287 \pm .056$ & $0.4307 \pm .100$ & $0.5514 \pm .035$ & $0.4255 \pm .041$ & $0.6535 \pm .067$ \\
        G-Blend On \citep{Wang_Tran_Feiszli_2020}
            & $0.2519 \pm .064$ & $0.4128 \pm .066$ & $0.5139 \pm .028$ & $0.4503 \pm .010$ & $0.5566 \pm .069$ \\
        G-Blend Off \citep{Wang_Tran_Feiszli_2020}
            & $0.2383 \pm .086$ & $0.4290 \pm .066$ & $0.5311 \pm .017$ & $0.4101 \pm .023$ & $0.6241 \pm .023$ \\
        \midrule 

        \ac{simbamm}$^{\text{CLS}}$
            & ${0.3229 \pm .054}$ & ${0.4936 \pm .015}$ & $0.5086 \pm .042$ & ${0.4351 \pm .014}$ & $0.6723 \pm .031$ \\
        \ac{simbamm}
            & $0.3056 \pm .080$ & $0.3950 \pm .101$ & ${0.5276 \pm .059}$ & $0.4326 \pm .028$ & n/a \\

        \arrayrulecolor{black}
        
        \bottomrule
    \end{tabular}
    
\end{table*}

\section{Detailed Efficiency Analysis}
\label{app:detailed_efficiency_analysis}
\begin{figure}[h]
    \centering
    \includegraphics[width=.6\linewidth]{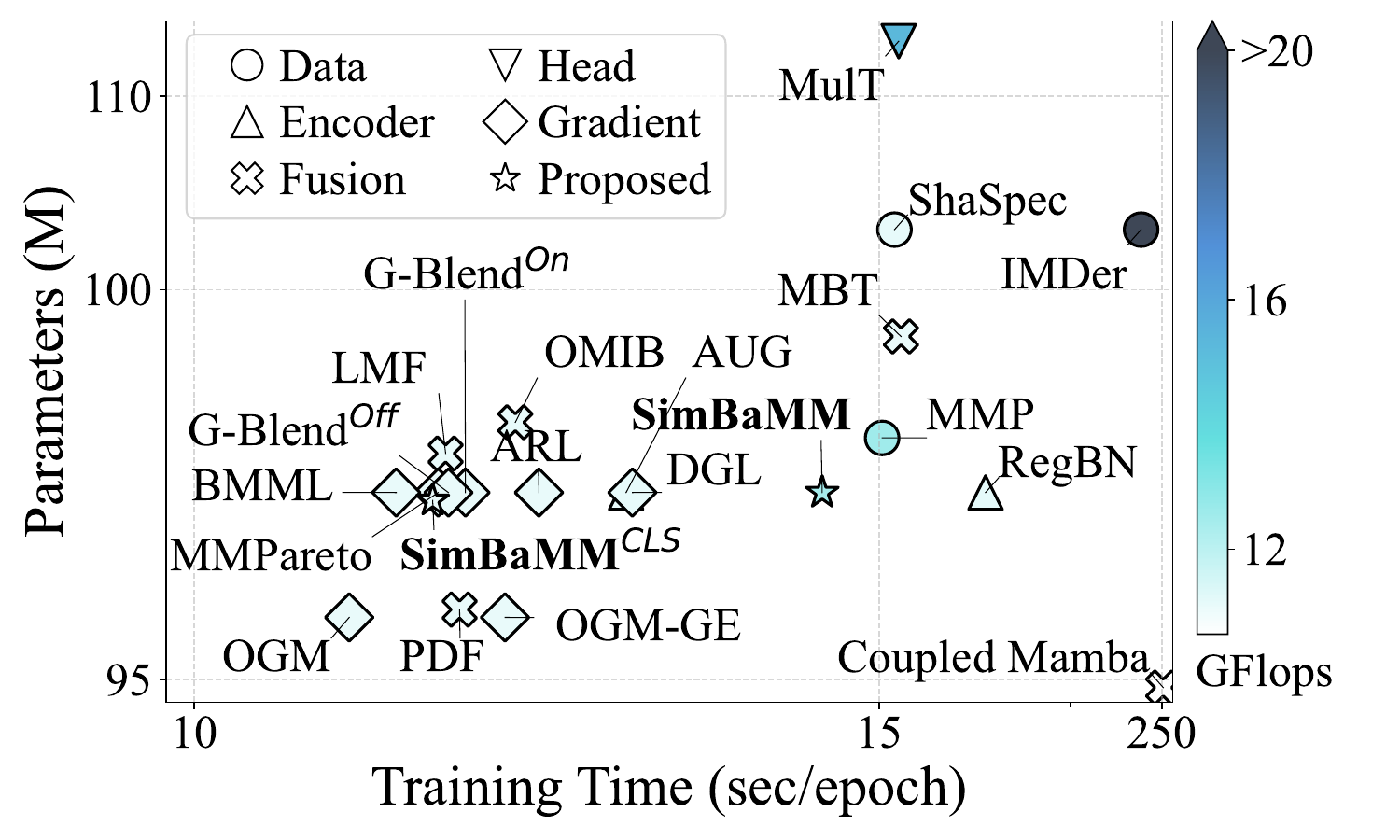}
    \caption{Visual representation of the efficiency of the methods w.r.t. the number of parameters, training time, and \acs{flops}.}
    \label{fig:efficiency_analysis}
\end{figure}

In addition to the tables representing efficiency vs. performance, \Cref{fig:efficiency_analysis,tab:results_params_flops} present a focused view on efficiency comparisons.
\begin{table}[h]
    \centering
    \small
    \caption{Tabular representation of the efficiency of the methods w.r.t. the number of parameters, training time, and \acs{flops}.}
    \label{tab:results_params_flops}

    \begin{tabular}{@{}c l cccc ccccc@{}}
    \toprule

    \textbf{Group} & 
    \textbf{Method} & 
    \textbf{\#Parameters (M) $\downarrow$} &
    {\textbf{FLOPs (G) $\downarrow$}} &
    {\textbf{Training Time (sec/epoch) $\downarrow$}}\\
    \midrule

    \arrayrulecolor{black!40}

    \multirow{3}{*}{Data}
    & IMDer \citep{Wang_Li_Cui_2023} 
        & 103.0 & 408.23 & 203.50 \\
    & MMP \citep{Nezakati_Reza_Patil_Solh_Asif_2024} 
        & 98.1 & 12.57 & 15.42 \\
    & ShaSpec \citep{Wang_Chen_Ma_Avery_Hull_Carneiro_2023} 
        & 103.0 & 11.14 & 17.45 \\

    \midrule

    \multirow{1}{*}{Encoder}
    & RegBN \citep{Ghahremani_Wachinger_2023} 
        & 97.4 & 11.15 & 43.21 \\
    & AUG \citep{Jiang_Huang_Yang_2025} 
        & 97.4 & 11.10 & 12.15 \\
    \midrule

    \multirow{6}{*}{Fusion}
    & MBT \citep{Nagrani_Yang_Arnab_Jansen_Schmid_Sun_2021} & 99.4 & 11.11 & 18.62 \\
    %& MCR \citep{Kontras_Strypsteen_Chatzichristos_Liang_Blaschko_Vos_2025} & 97.5 & 11.10 & 29.92 \\
    & LMF \citep{Liu_Shen_Lakshminarasimhan_Liang_Zadeh_Morency_2018} & 97.9 & 11.09 & 11.83 \\
    & PDF \citep{Cao_Xia_Ding_Zhang_Hu_2024} & 95.9 & 11.08 & 11.94 \\
    & Coupled Mamba \citep{Li_Zhou_Yu_Song_Yang_2024} & 94.9 & 10.63 & 253.80 \\
    & OMIB \citep{Wu_Shao_Wang_Sun_2025} & 98.3 & 11.11 & 12.34 \\
    \midrule

    \multirow{1}{*}{Head}
    & MulT \citep{Tsai_Bai_Liang_Kolter_Morency_Salakhutdinov_2019} 
        & 113.0 & 15.32 & 18.20 \\
    \midrule

    \multirow{9}{*}{Gradient}
    & OGM \citep{Peng_Wei_Deng_Wang_Hu_2022} 
        & 95.8 & 11.08 & 11.13 \\
    & $\text{OGM}^{\text{GE}}$ \citep{Peng_Wei_Deng_Wang_Hu_2022}
        & 95.8 & 11.08 & 12.27 \\
    & DGL \citep{Wei_Luo_Luo_2025a} 
        & 97.4 & 11.14 & 13.20 \\
    & ARL \citep{Wei_Luo_Luo_2025b} 
        & 97.4 & 11.10 & 12.52 \\
    & MMPareto \citep{Wei_Hu_2024} 
        & 97.4 & 11.10 & 11.78 \\
    & BMML \citep{Wu_Jastrzebski_Cho_Geras_2022}
        & 97.4 & 11.10 & 11.47 \\
    & $\text{G-Blend}^{\text{Online}}$ \citep{Wang_Tran_Feiszli_2020} 
        & 97.4 & 11.10 & 11.98 \\
    & $\text{G-Blend}^{\text{Offline}}$ \citep{Wang_Tran_Feiszli_2020} 
        & 97.4 & 11.10 & 11.85 \\
    \midrule

    \multirow{1}{*}{}
    & $\text{\ac{simbamm}}^{\text{CLS}}$
        & 97.3 & 11.10 & 11.74 \\
    & \ac{simbamm}
        & 97.4 & 12.46 & 14.58 \\

    \arrayrulecolor{black}
    \bottomrule
\end{tabular}
\end{table}

\section{Dataset Details}
\label{app:dataset_details}
\paragraph{MIMIC Symile}
    The MIMIC Symile dataset provides a challenging clinical benchmark for multi-label diagnostic prediction. Derived from the MIMIC database, it comprises \num{10345} samples collected from patients in intensive care units. The task is to predict the presence of ten conditions: Fracture, Enlarged Cardiomediastinum, Consolidation, Atelectasis, Edema, Cardiomegaly, Lung Lesion, Lung Opacity, Pneumonia, and Pneumothorax. Our evaluation leverages three distinct data types: laboratory test results, chest X-ray images, and \acp{ecg}. We use an \ac{mlp} for laboratory tests, a \ac{vit} for vision data, and a Transformer encoder for \acp{ecg}. The \ac{ecg} samples are preprocessed by removing baseline wander.

\paragraph{MIMIC HAIM}
    Similarly, the MIMIC HAIM dataset is a healthcare benchmark focused on the same ten-class diagnostic prediction task as Symile. This bimodal dataset is substantially larger, containing \num{45050} samples. For this dataset, our experiments utilize laboratory values and chest X-ray images as the two input modalities. We use an \ac{mlp} for laboratory tests and a \ac{vit} for vision data. 

\paragraph{INSPECT}
The INSPECT dataset \citep{Huang_Huo_Steinberg_Chiang_Lungren_Langlotz_Yeung_Shah_Fries_2023} comprises \num{22449} CT pulmonary angiography (CTPA) studies for pulmonary embolism (PE) diagnosis, combining 3D \ac{ct} images with longitudinal \acp{ehr}. We focus on the binary PE detection task, where approximately 20\% of cases are PE-positive. We use MOTOR \citep{Steinberg_Fries_Xu_Shah_2024} embeddings for structured EHR data and Med3dVLM \citep{Xin_Ates_Gong_Shao_2025} embeddings (latest vision encoder stage with global average pooling on the feature maps) for the CT volumes, in combination with \ac{mlp} encoders.

\begin{table}[ht]
    \centering
    \small
    \caption{
        Unimodal performance of the modalities for the healthcare datasets. For the \ac{ukb}, if unimodal data is not available for a sample, it is zero-imputed to ensure a fair comparison to the multimodal performance. 
    }
    \label{tab:unimodal_ukb_union}
    
    \begin{tabular}{@{}l c@{}}  
        \toprule
        
        \textbf{Modality} & 
        {\textbf{\acs{auroc}} $\uparrow$} \\ 
        
        \midrule

        \arrayrulecolor{black!40}

        \multicolumn{2}{l}{\hspace{-6pt}\textit{HAIM}} \\
        X-ray & $0.7042 \pm 0.020$ \\
        Laboratory & $0.6654 \pm 0.038$ \\
        \midrule 

        \multicolumn{2}{l}{\hspace{-6pt}\textit{Symile}} \\
        X-ray & $0.6453 \pm 0.030$ \\
        Laboratory & $0.5597 \pm 0.010$ \\
        \ac{ecg} & $0.4960 \pm 0.024$ \\
        \midrule 

        \multicolumn{2}{l}{\hspace{-6pt}\textit{INSPECT}} \\
        \ac{ct} & $0.5825 \pm 0.007$ \\
        \ac{ehr} & $0.6568 \pm 0.003$ \\
        \midrule 

        \multicolumn{2}{l}{\hspace{-6pt}\textit{UKB}} \\
        Metabolomics
            & $0.7108 \pm 0.003$ \\
        \ac{ehr} 
            & $0.7560 \pm 0.010$ \\
        Proteomics
            & $0.5434 \pm 0.009$ \\
        \Ac{prs}
            & $0.6008 \pm 0.008$ \\
        Blood biochemistry
            & $0.7156 \pm 0.004$ \\
        Baseline characteristics
            & $0.7151 \pm 0.004$ \\
        Local environment
            & $0.5049 \pm 0.009$ \\
        Arterial stiffness
            & $0.5290 \pm 0.016$ \\
        Anthropometry
            & $0.6086 \pm 0.025$ \\
        Blood pressure
            & $0.6403 \pm 0.008$ \\
        \ac{ecg}
            & $0.5298 \pm 0.017$ \\
        Eye measures
            & $0.5233 \pm 0.014$ \\
        Bone densitometry
            & $0.5853 \pm 0.003$ \\
        Grip strength
            & $0.5227 \pm 0.008$ \\
        Spirometry
            & $0.6668 \pm 0.021$ \\
        Touchscreen
            & $0.7334 \pm 0.007$ \\
        Cognitive function
            & $0.5956 \pm 0.006$ \\
        Hearing
            & $0.5355 \pm 0.020$ \\
        Interview
            & $0.6803 \pm 0.005$ \\
        Blood count
            & $0.6810 \pm 0.005$ \\
        Urine assays
            & $0.5384 \pm 0.006$ \\
        Telomeres
            & $0.5670 \pm 0.010$ \\
        Infectious diseases
            & $0.5008 \pm 0.001$ \\

        \arrayrulecolor{black}
        
        \bottomrule
    \end{tabular}
\end{table}

\paragraph{\ac{ukb}}
    The \ac{ukb} is a large, long-term prospective biobank study with \num{488131} samples after preprocessing and exclusions. While various targets\eg for survival analysis are possible, we chose to predict ten-year mortality from the recruitment date for simplicity. It comprises \num{23} modalities\ie metabolomics, \acs{ehr}, \acs{prs}, proteomics, blood biochemistry, environment, stiffness, anthropometry, blood pressure, \ac{ecg}, eye measures, bone densitometry, grip strength, spirometry, touchscreen, cognitive function, hearing tests, interviews, blood counts, urine assays, telomeres, and infectious diseases. The \ac{ukb} category IDs used for the structured field extraction are: 220 (NMR metabolomics), 1838 (Proteomics), 100008 (Anthropometry), 100007 (Arterial stiffness), 100094 (Baseline characteristics), 17518 (Blood biochemistry), 100081 (Blood count), 100011 (Blood pressure), 100018 (Bone densitometry of heel), 100026 (Cognitive function), 104 (ECG at rest), 100012 (ECG during exercise), 100013 (Eye measures), 300/264/100313 (Genetics), 100019 (Hand grip strength), 100049 (Hearing test), 51428 (Infectious diseases), 113 (Local environment), 100020 (Spirometry), 265 (Telomeres), 100025 (Touchscreen), 100083 (Urine assays), and 100071 (Verbal interview). We use \acp{mlp} for all modalities and raw inputs except for the \ac{ehr} modality, for which we use QWEN \citep{Zhang_Li_Long_Zhang_Lin_Yang_Xie_Yang_Liu_Lin_etal_2025} embeddings based on previously used serializations \citep{Hegselmann_Arnim_Rheude_Kronenberg_Sontag_Hindricks_Eils_Wild_2025}. Individual missing entries within a modality were imputed with zeros, and instances where a feature vector was entirely absent were excluded.

\begin{table}[ht]
    \centering
    \small
    \caption{
        Unimodal performance of the modalities for the \ac{ukb} without zero imputation\ie only the available modality subset is used. 
    }
    \label{tab:unimodal_ukb_intersect}
    
    \begin{tabular}{@{}l c@{}}  
        \toprule
        
        \textbf{Modality} & 
        {\textbf{\acs{auroc}} $\uparrow$} \\ 
        
        \midrule

        \arrayrulecolor{black!40}

        Metabolomics
            & $0.7118 \pm 0.004$ \\
        \ac{ehr} 
            & $0.7728 \pm 0.005$ \\
        Proteomics
            & $0.7588 \pm 0.018$ \\
        \Ac{prs}
            & $0.6009 \pm 0.009$ \\
        Blood biochemistry
            & $0.7183 \pm 0.004$ \\
        Baseline characteristics
            & $0.7150 \pm 0.006$ \\
        Local environment
            & $0.5025 \pm 0.004$ \\
        Arterial stiffness
            & $0.5997 \pm 0.009$ \\
        Anthropometry
            & $0.6060 \pm 0.013$ \\
        Blood pressure
            & $0.6388 \pm 0.009$ \\
        \ac{ecg}
            & $0.6869 \pm 0.005$ \\
        Eye measures
            & $0.5669 \pm 0.045$ \\
        Bone densitometry
            & $0.6008 \pm 0.002$ \\
        Grip strength
            & $0.5212 \pm 0.008$ \\
        Spirometry
            & $0.6615 \pm 0.004$ \\
        Touchscreen
            & $0.7180 \pm 0.006$ \\
        Cognitive function
            & $0.5956 \pm 0.005$ \\
        Hearing
            & $0.6020 \pm 0.009$ \\
        Interview
            & $0.6656 \pm 0.006$ \\
        Blood count
            & $0.6838 \pm 0.006$ \\
        Urine assays
            & $0.5284 \pm 0.009$ \\
        Telomeres
            & $0.5685 \pm 0.009$ \\
        Infectious diseases
            & $0.5330 \pm 0.027$ \\

        \arrayrulecolor{black}
        
        \bottomrule
    \end{tabular}
\end{table}

\paragraph{CMU-MOSI and CMU-MOSEI}
    Shifting from the clinical domain, the CMU-MOS(E)I dataset is a benchmark for multimodal sentiment analysis and emotion recognition. It consists of \num{2199} (MOSI) and \num{22856} (MOSEI) video clips of speakers expressing opinions, with annotations for seven emotional classes. The data is trimodal, encompassing visual, acoustic, and linguistic information. A key distinction from the other datasets is the use of the officially provided pre-computed feature embeddings, in contrast to processing raw data. We use BERT for language sequences and a Transformer encoder for the vision and audio embeddings. 

\paragraph{CH-SIMS and CH-SIMS 2}
    These datasets are benchmarks for multimodal sentiment analysis and emotion recognition. They consist of \num{2281} (CH-SIMS) and \num{4403} (CH-SIMS 2) video clips of speakers expressing emotions. Similar to MOS(E)I, they are trimodal but provide raw data. We use Wav2Vec2, ViViT, and BERT for the audio, video, and language data, respectively.

\paragraph{Crema-D}
    Crema-D is a dataset for multimodal sentiment analysis with six emotion classes (such as anger and fear) that can be further subdivided into a more fine-grained set of \num{24} classes. It consists of \num{7442} samples and provides audio and video data. Following related work, we use spectrograms in combination with a ResNet for the audio data and video frames in combination with another ResNet for the vision data. For the case study, we use horizontal flipping augmentations and ImageNet normalization values analogous to the AUG code repository. 

%\paragraph{VGGSound}
%    VGGSound is a dataset with $199,175$ samples for classification tasks of $309$ classes and video and audio data. Following related work, we use spectograms in combination with a ResNet for the audio data and extracted images in combination with another ResNet for the vision data. 
%    We create embeddings for being able to tune hyperparameters of the evaluated methods. We select pre-trained embedding models without supervised class-specific pre-training w.r.t. Crema-D and discard sophisticated tunings such as multi-clip evaluation. For the vision modality embeddings, we use V-JEPA2 \citep{Assran_Bardes_Fan_Garrido_Howes_Mojtaba_Komeili_Muckley_Rizvi_Roberts_etal2025} with $64$ frames per video and average-pooling of the resulting token sequence to a fixed temporal resolution of $256$ tokens.
%    For the audio modality embeddings, we use CLAP \citep{Elizalde_Deshmukh_Ismail_Wang_2023} by flattening the spatial-temporal output map into a sequence.

%\paragraph{Kinetics-400}
%    Kinetics is a robotics dataset with $200,892$ samples and $400$ (or optionally $600$ / $700$) classes. Since some video files are corrupted, $1,679$ samples are removed. We create embeddings for being able to tune hyperparameters of the evaluated methods. We select pre-trained embedding models without supervised class-specific pre-training w.r.t. Kinetics-400 and discard sophisticated tunings such as multi-clip evaluation. We follow the same embedding models as for VGGSound.

\section{Case Study Details}
\label{app:case_study_details}
Especially for the case study, alternative viewpoints on our set of implementation choices are possible. Therefore, we record key concerns, add further experiments and explain how they relate to the case study.

\paragraph{On the Necessity of Reusing Splits for Fair Comparison}
It is often argued that new methods should use legacy data splits to ensure numbers are directly comparable to previous tables. Comparability with prior results is a pragmatic concern, but we do not consider reusing legacy splits as a methodological requirement. In particular, when existing splits are underspecified or violate stated dataset guidance\eg subject independence, adopting a standardized, leakage-avoiding protocol is justified even if it breaks comparability with earlier reports. Rather than treating inherited splits as binding, we view them as one point in the design space: authors can report results on legacy splits for continuity while also providing results under a corrected protocol to support valid conclusions. In this sense, the case study targets evaluation practice\ie how conclusions shift under a more defensible protocol, rather than attributing shortcomings to any single method or paper.

%\paragraph{Prior Work on Crema-D}
%AUG follows the Crema-D splits established in prior work to enable direct comparison, which raises the question of whether a case study should instead focus on the earliest paper that introduced these splits. We do not view “earliest adoption” as the most informative choice for illustrating the impact of evaluation practice. Our goal is to show how widely used, currently influential protocols can shape reported conclusions. In that sense, recency and current prevalence in the literature are more relevant than historical provenance. Moreover, the community is not bound to legacy splits\ie any work can re-establish subject-independent splits under a standardized protocol, so tracing the protocol back to its first adopter is not necessary for the methodological point. Finally, our evaluation includes OGM (a canonical method using these splits) alongside AUG, ensuring that the effect of the commonly used protocol is represented in the broader comparison rather than being attributed to a single method in isolation.

\begin{figure}[t]
    \centering
    \includegraphics[width=.6\linewidth]{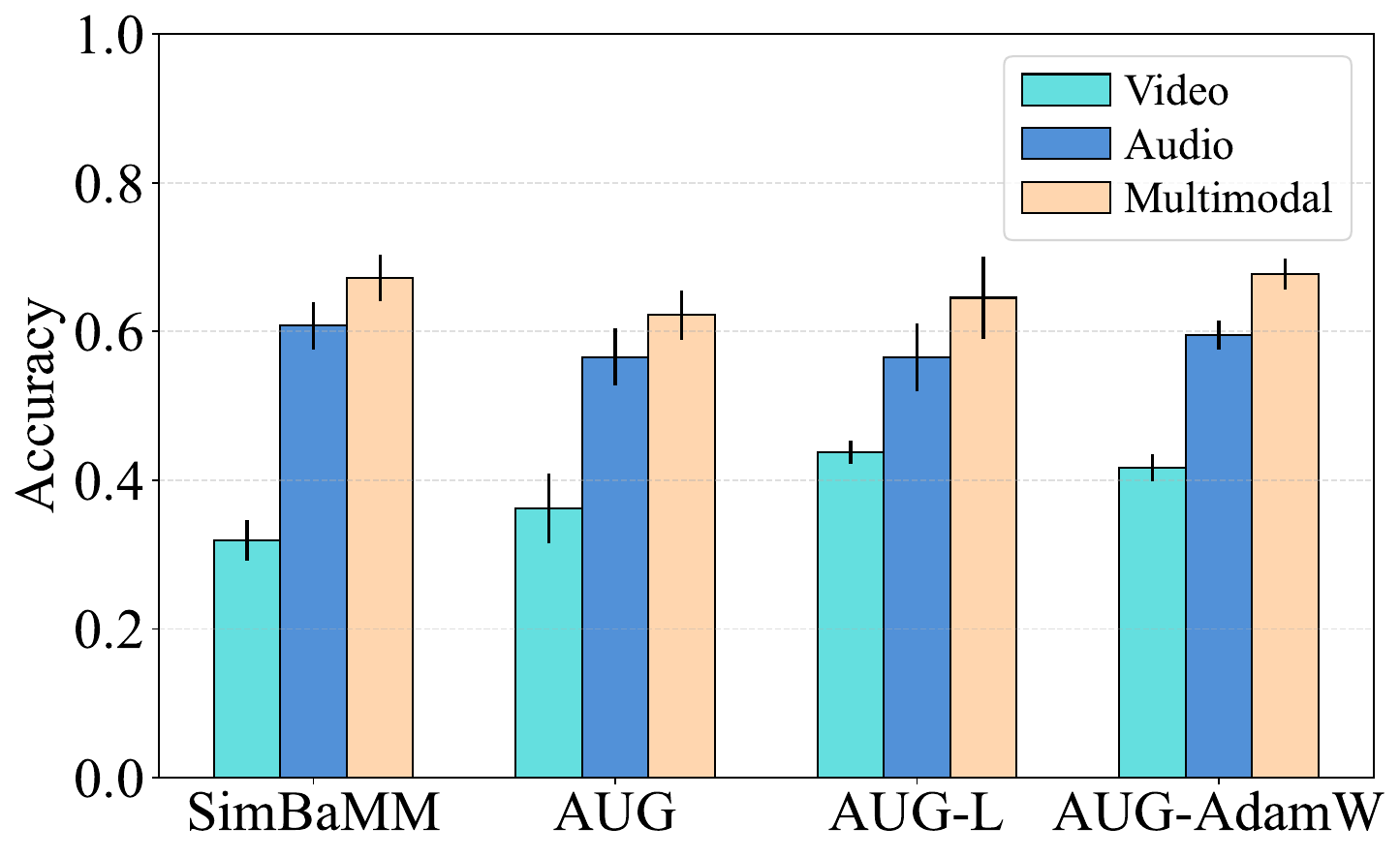}
    \caption{Alternative configurations for AUG in the case study setting with the corrected data protocol\ie with a linear head (AUG-L) and another optimizer (AUG-AdamW).}
    \label{fig:case_study_details}
\end{figure}

\paragraph{AUG with Linear Head}
We replace AUG’s original linear head with a Transformer-based head to align the architectural capacity with other compared methods. To verify that this choice does not drive our conclusions, we additionally show that AUG with a linear head performs similarly (AUG-L, \Cref{fig:case_study_details}). AUG shows the same overall performance trends and leads to the same conclusions about its relative behavior in the multimodal setting, indicating that the case study findings are not an artifact of the Transformer head.

\paragraph{Different Optimizers}
To assess whether our conclusions depend on the choice of optimizer, we performed an additional hyperparameter sweep using AdamW instead of the originally used SGD (AUG-AdamW, \Cref{fig:case_study_details}). To circumvent learning rate schedulers, we still use the ScheduleFree variant. This change does not alter the qualitative outcome: the relative behavior of AUG and the baselines, as well as the resulting ranking trends that motivate the case study, remain consistent. This indicates that our headline conclusions are not an artifact of selecting a particular optimizer family, but persist under a substantially different optimization setup.

\paragraph{Sweep on Full Parameter Space}
One may argue that initializing AUG’s sweep from architectural choices selected for our method\eg encoder and head dimensions, could bias the comparison, and that AUG should instead be tuned from scratch over the full design space, including architectural hyperparameters as well as optimization settings. This concern is addressed by AUG-L (\Cref{fig:case_study_details}), since all model hyperparameters are sweeped in this setting, except for the encoders which are set to the originally used ResNet18s. The resulting performance and qualitative trends are unchanged: the relative conclusions of the case study and the observed method ranking remain consistent. This indicates that our findings are robust to the hyperparameter search procedure and do not arise from anchoring AUG to architectural choices optimized for a different method.

\paragraph{Differences to Original Results}
For DGL and ARL on CREMA-D, our experiments reveal the same issue identified for AUG: the reported gains rely on model selection using the test set. When applying our strictly separated validation protocol, performance decreases substantially. We reproduced this in the official code repositories as well. 
First, as for our case study (on AUG), the train/val/test splits contain overlapping speaker identities across training and evaluation sets. Second, checkpoints are selected using test-set performance.
We replaced these with speaker-disjoint $60/20/20$ train/validation/test splits and selected checkpoints on the validation set. With these corrections, the official implementations reach decreased performance comparable to the results which we report in our paper. For DGL, this means the original result of $77.48$ decreases to $67.3$. For ARL, this means that the original result of $76.61$ decreases to $65.9$

\section{Statistical Comparison Across Datasets}
\label{app:statistical_comparison}
To rigorously compare methods across datasets, we employ Bayesian hierarchical analysis following \citet{Benavoli_Corani_Demsar_Zaffalon_2017,Corani_Benavoli_Demsar_Mangili_Zaffalon_2017}. The model uses fold-level cross-validation results, accounts for correlation between folds due to overlapping training sets, and yields direct posterior probabilities for each pairwise comparison.

\subsection{Bayesian Hierarchical Pairwise Comparisons}
\label{app:statistical_comparison_omnibus}
For each pair of methods $(i,j)$, we compare their fold-level cross-validation results across the datasets on which both methods are available. Let $x_{d,f}=s_{d,i,f}-s_{d,j,f}$ be the per-fold performance difference on dataset $d$. We model within-dataset differences as correlated observations due to shared training data in $K$-fold cross-validation:
\begin{equation}
    \mathbf{x}_d \sim \mathcal{N}(\mu_d \mathbf{1}, \Sigma_d), \qquad \rho = 1/K,
\end{equation}
where $K$ is the number of folds (here $K=5$, thus $\rho=0.2$). Dataset-level mean differences are modeled with a robust population distribution:
\begin{equation}
    \mu_d \sim t(\delta_0, \sigma_0, \nu),
\end{equation}
where $\delta_0$ is the population-level mean difference and $\nu$ represents the degrees of freedom, accounting for non-normality. We define a \ac{rope} of $\pm \epsilon$ (here $\epsilon=0.01$ in absolute metric points). Following \citet{Corani_Benavoli_Demsar_Mangili_Zaffalon_2017}, we report the posterior probabilities that each of three outcomes is the \emph{most probable} outcome on a new dataset drawn from the population distribution $\delta_{\text{new}} \sim t(\delta_0, \sigma_0, \nu)$: \emph{method $j$ better} ($\delta_{\text{new}}<-\epsilon$), \emph{practically equivalent} ($|\delta_{\text{new}}|\le\epsilon$), and \emph{method $i$ better} ($\delta_{\text{new}}>\epsilon$). Concretely, for each posterior draw of $(\delta_0,\sigma_0,\nu)$ we compute the three region probabilities for $\delta_{\text{new}}$ and count which region has the largest probability; the reported values are the resulting frequencies.
We complete the model with the following hyperpriors based on the robust ranges identified in \citet{Corani_Benavoli_Demsar_Mangili_Zaffalon_2017}:
\begin{align}
    \delta_0 &\sim \mathcal{U}(-1, 1), \\
    \sigma_0 &\sim \mathcal{U}(0, \bar{s}_0), \\
    \nu &\sim \text{Gamma}(\alpha, \beta) + 1, \\
    \alpha &\sim \mathcal{U}(1.0, 2.0), \quad \beta \sim \mathcal{U}(0.01, 0.10).
\end{align}
Following \citet{Corani_Benavoli_Demsar_Mangili_Zaffalon_2017}, we set $\bar{s}_0 = 1000\,s_{\bar{x}}$, where $s_{\bar{x}}$ is the standard deviation across datasets of the per-dataset mean differences $\bar{x}_d=\frac{1}{K}\sum_{f=1}^{K} x_{d,f}$.

\paragraph{Reported quantities} For every comparison, we additionally report the posterior mean and 95\% credible interval of the population-level mean difference $\delta_0$ (absolute metric points; $\delta_0>0$ indicates method $i$ is better on average).

\paragraph{Handling missingness} Missing method-dataset pairs are handled via pairwise complete cases: for a comparison $(i,j)$ we include only datasets where \emph{both} methods have fold-level results available.

\paragraph{Model fitting and diagnostics} We fit the model with NUTS using PyMC (nutpie backend) with four chains (2000 warmup and 4000 post-warmup draws per chain). We monitor $\hat{R}$, effective sample size (ESS), divergences, and max tree depth hits for the key parameters $(\delta_0,\sigma_0,\nu)$ to ensure reliable posterior estimates. When diagnostics are poor, we automatically re-sample with a more conservative target acceptance rate and/or a larger draw budget. \Cref{tab:bayesian_diagnostics} summarizes these diagnostics for the baseline comparisons reported in \Cref{tab:bayesian_simbamm_cls,tab:bayesian_unimodal}.

\paragraph{Bounded-metric sanity check} Since the evaluated metrics are bounded in $[0,1]$, the dataset-level mean difference satisfies $\delta \in [-1,1]$. We therefore compute a simple posterior predictive check: the probability that a new dataset-level effect drawn from the fitted population distribution violates this bound, i.e., $P(|\delta_{\text{new}}|>1)$. For the baseline comparisons in \Cref{tab:bayesian_diagnostics}, this probability is negligible, indicating the fitted population model places essentially no mass outside the feasible range.

\begin{table}[t]
    \centering
    \footnotesize
    \setlength{\tabcolsep}{1.5pt}
    \caption{Comparison with the best unimodal baseline.}
    \label{tab:bayesian_unimodal}
    \begin{tabular}{lccccc}
    \toprule
    \textbf{Method} & \makecell{\textbf{P(base}\\\textbf{$>$other)}} & \makecell{\textbf{P($\approx$)}} & \makecell{\textbf{P(other}\\\textbf{$>$base)}} & \makecell{\textbf{$\mathbb{E}[\delta_0]$}} & \makecell{\textbf{95\% CI}} \\
    \midrule
    CoupledMamba & 0.18 & 0.10 & 0.72 & -0.010 & [-0.038,0.024] \\
    SimBaMM-CLS & 0.41 & 0.06 & 0.53 & -0.002 & [-0.031,0.031] \\
    MMPareto & 0.66 & 0.08 & 0.26 & 0.013 & [-0.026,0.066] \\
    ARL & 0.19 & 0.14 & 0.67 & -0.008 & [-0.034,0.024] \\
    GBlend-On & 0.61 & 0.10 & 0.29 & 0.011 & [-0.029,0.062] \\
    BMML & 0.59 & 0.03 & 0.38 & 0.006 & [-0.035,0.056] \\
    OGM-GE & 0.80 & 0.02 & 0.18 & 0.017 & [-0.024,0.060] \\
    MulT & 0.29 & 0.67 & 0.04 & 0.006 & [-0.008,0.025] \\
    SimBaMM-Full & 0.74 & 0.14 & 0.13 & 0.021 & [-0.023,0.083] \\
    MBT & 0.91 & 0.03 & 0.06 & 0.021 & [-0.009,0.054] \\
    GBlend-Off & 0.73 & 0.18 & 0.08 & 0.015 & [-0.014,0.046] \\
    LMF & 0.93 & 0.00 & 0.07 & 0.057 & [-0.020,0.142] \\
    OMIB & 0.96 & 0.00 & 0.04 & 0.033 & [-0.007,0.075] \\
    OGM & 0.80 & 0.05 & 0.14 & 0.018 & [-0.020,0.064] \\
    PDF & 0.95 & 0.03 & 0.03 & 0.023 & [-0.002,0.050] \\
    AUG & 0.94 & 0.02 & 0.04 & 0.031 & [-0.005,0.072] \\
    RegBN & 0.68 & 0.04 & 0.27 & 0.013 & [-0.026,0.067] \\
    DGL & 0.98 & 0.00 & 0.02 & 0.083 & [0.009,0.161] \\
    \bottomrule
    \end{tabular}
\end{table}

\subsection{\ac{rope} Sensitivity}
To assess how sensitive our conclusions are to the practical significance threshold, we repeat the comparisons with $\epsilon \in \{0.005, 0.01, 0.02\}$ (i.e., \ac{rope} $\pm$0.5\%, $\pm$1\%, and $\pm$2\% in absolute metric points). \Cref{tab:bayesian_simbamm_cls,tab:bayesian_unimodal} correspond to $\epsilon=0.01$; additional \ac{rope} settings are reported below.

\paragraph{Key takeaway} The qualitative conclusions reported in \Cref{subsec:downstream_task_analysis} are robust to reasonable \ac{rope} choices: tightening the \ac{rope} makes more comparisons count as ``wins'' by a small margin (reducing $P(\approx)$), whereas widening the \ac{rope} shifts probability toward practical equivalence. Importantly, the posterior on $\delta_0$ (mean and 95\% credible interval) is unchanged; only the partitioning into \emph{win / \ac{rope} / loss} depends on $\epsilon$.

\paragraph{Comparison to SimBaMM$^{\textbf{CLS}}$} For the strongest competitors, the interpretation changes smoothly with $\epsilon$. For example, CoupledMamba has $P(\text{CoupledMamba}>\text{SimBaMM-CLS})=0.64$ under $\epsilon=0.005$ (\Cref{tab:bayesian_simbamm_cls_rope005}), but this drops to $0.30$ under $\epsilon=0.01$ (\Cref{tab:bayesian_simbamm_cls}) and to $0.04$ under $\epsilon=0.02$ (\Cref{tab:bayesian_simbamm_cls_rope020}), while $P(\approx)$ increases from $0.28 \rightarrow 0.67 \rightarrow 0.95$. Thus, if a $\geq 1\%$ improvement is considered practically meaningful (as in \Cref{subsec:downstream_task_analysis}), no method shows strong evidence of outperforming SimBaMM-CLS; if a tighter threshold is used, CoupledMamba appears more likely to be \emph{slightly} better, consistent with a small estimated mean effect (\Cref{tab:bayesian_simbamm_cls} reports $\mathbb{E}[\delta_0]\approx -0.006$ for SimBaMM-CLS minus CoupledMamba).

\paragraph{Comparison to the best unimodal baseline} The sensitivity analysis suggests that improvements over the best unimodal baseline (when present) are typically modest: CoupledMamba and ARL are likely to beat the unimodal baseline for $\epsilon\in\{0.005,0.01\}$ (e.g., $P(\text{CoupledMamba}>\text{Unimodal})=0.79/0.72$ and $P(\text{ARL}>\text{Unimodal})=0.76/0.67$ for $\epsilon=0.005/0.01$), but these probabilities decrease to $0.37$ and $0.28$ at $\epsilon=0.02$ with $P(\approx)$ becoming dominant (\Cref{tab:bayesian_unimodal_rope020}). This supports the interpretation in \Cref{subsec:downstream_task_analysis} that any gains over unimodal baselines are not reliably large in absolute terms.

\begin{table}[t]
    \centering
    \footnotesize
    \setlength{\tabcolsep}{1.5pt}
    \caption{
        Comparison with SimBaMM-CLS (\ac{rope} $\pm0.5\%$).
    }
    \label{tab:bayesian_simbamm_cls_rope005}
    \begin{tabular}{lccccc}
    \toprule
    \textbf{Method} & \makecell{\textbf{P(base}\\\textbf{$>$other)}} & \makecell{\textbf{P($\approx$)}} & \makecell{\textbf{P(other}\\\textbf{$>$base)}} & \makecell{\textbf{$\mathbb{E}[\delta_0]$}} & \makecell{\textbf{95\% CI}} \\
    \midrule
    CoupledMamba & 0.07 & 0.28 & 0.64 & -0.006 & [-0.021,0.006] \\
    Best unimodal & 0.56 & 0.00 & 0.44 & 0.002 & [-0.031,0.031] \\
    MMPareto & 0.23 & 0.32 & 0.45 & -0.002 & [-0.019,0.022] \\
    ARL & 0.45 & 0.18 & 0.37 & 0.002 & [-0.019,0.028] \\
    GBlend-On & 0.48 & 0.23 & 0.29 & 0.005 & [-0.018,0.045] \\
    BMML & 0.79 & 0.00 & 0.21 & 0.013 & [-0.021,0.051] \\
    OGM-GE & 0.89 & 0.03 & 0.08 & 0.015 & [-0.009,0.042] \\
    MulT & 0.47 & 0.13 & 0.41 & 0.002 & [-0.023,0.031] \\
    SimBaMM-Full & 0.62 & 0.07 & 0.31 & 0.007 & [-0.038,0.056] \\
    MBT & 0.90 & 0.05 & 0.05 & 0.012 & [-0.004,0.028] \\
    GBlend-Off & 0.95 & 0.01 & 0.05 & 0.021 & [-0.005,0.048] \\
    LMF & 0.90 & 0.00 & 0.10 & 0.049 & [-0.028,0.142] \\
    OMIB & 0.98 & 0.00 & 0.02 & 0.022 & [0.003,0.043] \\
    OGM & 0.95 & 0.01 & 0.04 & 0.017 & [-0.003,0.040] \\
    PDF & 0.99 & 0.00 & 0.01 & 0.020 & [0.005,0.037] \\
    AUG & 0.97 & 0.00 & 0.03 & 0.032 & [-0.000,0.066] \\
    RegBN & 0.92 & 0.03 & 0.05 & 0.014 & [-0.004,0.039] \\
    DGL & 0.98 & 0.00 & 0.02 & 0.078 & [0.006,0.158] \\
    \bottomrule
    \end{tabular}
\end{table}

\begin{table}[t]
    \centering
    \footnotesize
    \setlength{\tabcolsep}{1.5pt}
    \caption{
        Comparison with SimBaMM-CLS (\ac{rope} $\pm2\%$).
    }
    \label{tab:bayesian_simbamm_cls_rope020}
    \begin{tabular}{lccccc}
    \toprule
    \textbf{Method} & \makecell{\textbf{P(base}\\\textbf{$>$other)}} & \makecell{\textbf{P($\approx$)}} & \makecell{\textbf{P(other}\\\textbf{$>$base)}} & \makecell{\textbf{$\mathbb{E}[\delta_0]$}} & \makecell{\textbf{95\% CI}} \\
    \midrule
    CoupledMamba & 0.01 & 0.95 & 0.04 & -0.006 & [-0.021,0.006] \\
    Best unimodal & 0.25 & 0.53 & 0.22 & 0.002 & [-0.031,0.031] \\
    MMPareto & 0.05 & 0.91 & 0.04 & -0.002 & [-0.019,0.022] \\
    ARL & 0.11 & 0.84 & 0.06 & 0.002 & [-0.019,0.028] \\
    GBlend-On & 0.21 & 0.74 & 0.05 & 0.005 & [-0.018,0.045] \\
    BMML & 0.51 & 0.38 & 0.11 & 0.013 & [-0.021,0.051] \\
    OGM-GE & 0.38 & 0.59 & 0.03 & 0.015 & [-0.009,0.042] \\
    MulT & 0.13 & 0.79 & 0.08 & 0.002 & [-0.023,0.031] \\
    SimBaMM-Full & 0.30 & 0.57 & 0.14 & 0.007 & [-0.038,0.056] \\
    MBT & 0.14 & 0.85 & 0.01 & 0.012 & [-0.004,0.028] \\
    GBlend-Off & 0.61 & 0.38 & 0.01 & 0.021 & [-0.005,0.048] \\
    LMF & 0.87 & 0.04 & 0.10 & 0.049 & [-0.028,0.142] \\
    OMIB & 0.65 & 0.35 & 0.01 & 0.022 & [0.003,0.043] \\
    OGM & 0.37 & 0.62 & 0.01 & 0.017 & [-0.003,0.040] \\
    PDF & 0.46 & 0.54 & 0.00 & 0.020 & [0.005,0.037] \\
    AUG & 0.88 & 0.11 & 0.02 & 0.032 & [-0.000,0.066] \\
    RegBN & 0.29 & 0.71 & 0.01 & 0.014 & [-0.004,0.039] \\
    DGL & 0.98 & 0.00 & 0.02 & 0.078 & [0.006,0.158] \\
    \bottomrule
    \end{tabular}
\end{table}

\begin{table}[t]
    \centering
    \footnotesize
    \setlength{\tabcolsep}{1.5pt}
    \caption{
        Comparison with the best unimodal baseline (\ac{rope} $\pm0.5\%$).
    }
    \label{tab:bayesian_unimodal_rope005}
    \begin{tabular}{lccccc}
    \toprule
    \textbf{Method} & \makecell{\textbf{P(base}\\\textbf{$>$other)}} & \makecell{\textbf{P($\approx$)}} & \makecell{\textbf{P(other}\\\textbf{$>$base)}} & \makecell{\textbf{$\mathbb{E}[\delta_0]$}} & \makecell{\textbf{95\% CI}} \\
    \midrule
    CoupledMamba & 0.20 & 0.02 & 0.79 & -0.010 & [-0.038,0.024] \\
    SimBaMM-CLS & 0.44 & 0.00 & 0.56 & -0.002 & [-0.031,0.031] \\
    MMPareto & 0.69 & 0.02 & 0.29 & 0.013 & [-0.026,0.066] \\
    ARL & 0.23 & 0.02 & 0.76 & -0.008 & [-0.034,0.024] \\
    GBlend-On & 0.65 & 0.02 & 0.33 & 0.011 & [-0.029,0.062] \\
    BMML & 0.60 & 0.00 & 0.40 & 0.006 & [-0.035,0.056] \\
    OGM-GE & 0.81 & 0.00 & 0.19 & 0.017 & [-0.024,0.060] \\
    MulT & 0.63 & 0.28 & 0.09 & 0.006 & [-0.008,0.025] \\
    SimBaMM-Full & 0.80 & 0.04 & 0.16 & 0.021 & [-0.023,0.083] \\
    MBT & 0.94 & 0.00 & 0.06 & 0.021 & [-0.009,0.054] \\
    GBlend-Off & 0.84 & 0.04 & 0.12 & 0.015 & [-0.014,0.046] \\
    LMF & 0.93 & 0.00 & 0.07 & 0.057 & [-0.020,0.142] \\
    OMIB & 0.96 & 0.00 & 0.04 & 0.033 & [-0.007,0.075] \\
    OGM & 0.84 & 0.01 & 0.16 & 0.018 & [-0.020,0.064] \\
    PDF & 0.97 & 0.00 & 0.03 & 0.023 & [-0.002,0.050] \\
    AUG & 0.95 & 0.01 & 0.04 & 0.031 & [-0.005,0.072] \\
    RegBN & 0.71 & 0.00 & 0.29 & 0.013 & [-0.026,0.067] \\
    DGL & 0.98 & 0.00 & 0.02 & 0.083 & [0.009,0.161] \\
    \bottomrule
    \end{tabular}
\end{table}

\begin{table}[t]
    \centering
    \footnotesize
    \setlength{\tabcolsep}{1.5pt}
    \caption{
        Comparison with the best unimodal baseline (\ac{rope} $\pm2\%$).
    }
    \label{tab:bayesian_unimodal_rope020}
    \begin{tabular}{lccccc}
    \toprule
    \textbf{Method} & \makecell{\textbf{P(base}\\\textbf{$>$other)}} & \makecell{\textbf{P($\approx$)}} & \makecell{\textbf{P(other}\\\textbf{$>$base)}} & \makecell{\textbf{$\mathbb{E}[\delta_0]$}} & \makecell{\textbf{95\% CI}} \\
    \midrule
    CoupledMamba & 0.10 & 0.52 & 0.37 & -0.010 & [-0.038,0.024] \\
    SimBaMM-CLS & 0.22 & 0.53 & 0.25 & -0.002 & [-0.031,0.031] \\
    MMPareto & 0.52 & 0.34 & 0.14 & 0.013 & [-0.026,0.066] \\
    ARL & 0.09 & 0.63 & 0.28 & -0.008 & [-0.034,0.024] \\
    GBlend-On & 0.46 & 0.38 & 0.16 & 0.011 & [-0.029,0.062] \\
    BMML & 0.43 & 0.34 & 0.23 & 0.006 & [-0.035,0.056] \\
    OGM-GE & 0.66 & 0.22 & 0.12 & 0.017 & [-0.024,0.060] \\
    MulT & 0.07 & 0.92 & 0.01 & 0.006 & [-0.008,0.025] \\
    SimBaMM-Full & 0.54 & 0.39 & 0.07 & 0.021 & [-0.023,0.083] \\
    MBT & 0.61 & 0.36 & 0.03 & 0.021 & [-0.009,0.054] \\
    GBlend-Off & 0.40 & 0.56 & 0.03 & 0.015 & [-0.014,0.046] \\
    LMF & 0.93 & 0.01 & 0.07 & 0.057 & [-0.020,0.142] \\
    OMIB & 0.90 & 0.06 & 0.04 & 0.033 & [-0.007,0.075] \\
    OGM & 0.58 & 0.33 & 0.08 & 0.018 & [-0.020,0.064] \\
    PDF & 0.68 & 0.31 & 0.01 & 0.023 & [-0.002,0.050] \\
    AUG & 0.83 & 0.15 & 0.02 & 0.031 & [-0.005,0.072] \\
    RegBN & 0.51 & 0.35 & 0.15 & 0.013 & [-0.026,0.067] \\
    DGL & 0.98 & 0.00 & 0.02 & 0.083 & [0.009,0.161] \\
    \bottomrule
    \end{tabular}
\end{table}

\subsection{$\rho$ and Prior-Bound Sensitivity}
We follow the correlation heuristic $\rho=1/K$ and the weakly-informative uniform scale priors recommended by \citet{Corani_Benavoli_Demsar_Mangili_Zaffalon_2017}. To ensure conclusions are not driven by these modeling choices, we additionally repeat the baseline comparisons using (i) an alternative fold-correlation heuristic $\rho=1/(K-1)$ and (ii) tighter upper bounds for the uniform scale priors. \Cref{tab:bayesian_sensitivity} reports the maximum absolute changes in the reported probabilities and $\mathbb{E}[\delta_0]$ under these perturbations.

\begin{table}[t]
\centering
\scriptsize
\setlength{\tabcolsep}{2pt}
\caption{Sensitivity of baseline comparisons to $\rho$ and prior bounds.}
\label{tab:bayesian_sensitivity}
\begin{tabular}{llcc}
\toprule
\textbf{Baseline} & \textbf{Metric} & \textbf{$\rho=1/(K-1)$} & \textbf{$\bar{s}\,\times 100$} \\
\midrule
SimBaMM-CLS & \makecell[l]{max $|\Delta P|$\\(win)} & 0.034 & 0.015 \\
 & \makecell[l]{max $|\Delta P|$\\($\approx$)} & 0.053 & 0.024 \\
 & \makecell[l]{max $|\Delta P|$\\(loss)} & 0.024 & 0.024 \\
 & max $|\Delta \mathbb{E}[\delta_0]|$ & 0.001 & 0.001 \\
\midrule
Best unimodal & \makecell[l]{max $|\Delta P|$\\(win)} & 0.036 & 0.012 \\
 & \makecell[l]{max $|\Delta P|$\\($\approx$)} & 0.035 & 0.008 \\
 & \makecell[l]{max $|\Delta P|$\\(loss)} & 0.023 & 0.010 \\
 & max $|\Delta \mathbb{E}[\delta_0]|$ & 0.002 & 0.001 \\
\bottomrule
\end{tabular}
\end{table}

\paragraph{Sensitivity result} The sensitivity results show that conclusions are stable: across the baseline comparisons, the win/\ac{rope}/loss probabilities change by at most \num{0.053} and the population-level mean effect estimates $\mathbb{E}[\delta_0]$ shift by at most \num{0.002} under these alternative settings (\Cref{tab:bayesian_sensitivity}).

\paragraph{Sampler diagnostics} We summarize standard MCMC diagnostics for the baseline comparisons (\Cref{tab:bayesian_diagnostics}): the maximum split-$\hat{R}$ across the key parameters $(\delta_0,\sigma_0,\nu)$, the minimum bulk and tail effective sample sizes (ESS), and whether any fits exhibit NUTS divergences or max-treedepth hits. We also report the maximum posterior predictive probability $P(|\delta_{\text{new}}|>1)$ as a bounded-metric sanity check.

\begin{table}[t]
\centering
\scriptsize
\setlength{\tabcolsep}{2pt}
\caption{Sampler diagnostics for baseline comparisons.}
\label{tab:bayesian_diagnostics}
\begin{tabular}{lcccc}
\toprule
\textbf{Baseline} & \makecell{\textbf{max}\\\textbf{$\hat{R}$}} & \makecell{\textbf{min ESS}\\\textbf{(bulk/tail)}} & \makecell{\textbf{div}\\\textbf{/td}} & \makecell{\textbf{max}\\\textbf{$P(|\delta_{\text{new}}|>1)$}} \\
\midrule
SimBaMM-CLS & 1.010 & 439/108 & 0/0 & 7.07e-04 \\
Best unimodal & 1.010 & 501/320 & 0/0 & 3.86e-04 \\
\bottomrule
\end{tabular}
\end{table}

\paragraph{Diagnostics result} The sampler diagnostics indicate adequate convergence and effective sampling for the baseline comparisons: split-$\hat{R}$ is close to \num{1} (max \num{1.01}), ESS values are above common thresholds (min bulk/tail ESS \num{439}/\num{108}), and we observe no divergences or max-treedepth issues (\Cref{tab:bayesian_diagnostics}).

\section{Additional \ac{ukb} Experiments}
\label{app:ukb_m8_m16}

We include results on the \ac{ukb} with intermediate modality counts in \Cref{tab:results_ukb_m8_16}. Our results suggest that data scale may play an important role in realizing the benefits of multimodal learning. Multimodal models on the \ac{ukb} show gains over unimodal baselines, and additional experiments with eight (PRS, Local environment, ECG, Spirometry, Bone densitometry, Cognitive function, Eye measures, Urine assays) and $16$ (M=8 and Metabolomics, Blood pressure, Anthropometry, Hearing, Interview, Blood count, Telomeres, Infectious diseases) modalities further illustrate how performance trends evolve as the number of modalities increases. For MulT, computational cost scales quadratically with the number of modalities due to pairwise cross-modal attention. Hyperparameter tuning was therefore computationally infeasible for the 8- and 16-modality settings, for which we report n/a.
\begin{table}[h]
    \centering
    \small
    \caption{
        Performance for eight and $16$ modalities on the 
        \ac{ukb}. 
    }
    \label{tab:results_ukb_m8_16}
    
    \begin{tabular}{@{}l c c@{}}  
        \toprule
        
        \textbf{Method} & 
        {M8 \textbf{AUROC} $\uparrow$} &
        {M16 \textbf{AUROC} $\uparrow$} \\ 
        %{\textbf{VGG} $\uparrow$} &
        %{\textbf{Kinetics} $\uparrow$} \\
        \midrule

        \arrayrulecolor{black!40}

        Best Unimodal
            & $0.6869 \pm 0.005$ (ECG) 
            & $0.7118 \pm 0.004$ (Metabolomics) \\
        \midrule 

        \multicolumn{1}{l}{\hspace{-6pt}\textit{Missing Data-Based}} \\
        IMDer \citep{Wang_Li_Cui_2023} 
            & $0.7187 \pm 0.003$ 
            & $0.7689 \pm 0.004$ \\
        MMP \citep{Nezakati_Reza_Patil_Solh_Asif_2024}
            & $0.7203 \pm 0.001$ 
            & $0.7716 \pm 0.002$ \\
        ShaSpec \citep{Wang_Chen_Ma_Avery_Hull_Carneiro_2023}
            & $0.6831 \pm 0.015$ 
            & $0.7640 \pm 0.013$ \\
        
        \cmidrule(lr){2-3}
        \ac{simbamm}$^{\text{CLS}}$
            & $0.7211 \pm 0.003$ 
            & $0.7660 \pm 0.003$ \\
        \ac{simbamm} 
            & n/a 
            & n/a \\
        \midrule

        \multicolumn{1}{l}{\hspace{-6pt}\textit{Encoder-Based}} \\
        RegBN \citep{Ghahremani_Wachinger_2023} 
            & $0.7180 \pm 0.008$ 
            & $0.7661 \pm 0.002$ \\
        AUG \citep{Jiang_Huang_Yang_2025}
            & $0.6572 \pm 0.010$ 
            & $0.7582 \pm 0.007$ \\
        \midrule

        \multicolumn{1}{l}{\hspace{-6pt}\textit{Fusion-Based}} \\
        MBT \citep{Nagrani_Yang_Arnab_Jansen_Schmid_Sun_2021}
            & $0.7180 \pm 0.008$ 
            & $0.7511 \pm 0.005$ \\
        LMF \citep{Liu_Shen_Lakshminarasimhan_Liang_Zadeh_Morency_2018}
            & $0.5069 \pm 0.004$ 
            & $0.4997 \pm 0.001$ \\
        PDF \citep{Cao_Xia_Ding_Zhang_Hu_2024}
            & $0.7033 \pm 0.005$ 
            & $0.7646 \pm 0.004$ \\
        Coupled Mamba \citep{Li_Zhou_Yu_Song_Yang_2024}
            & $0.7182 \pm 0.003$ 
            & $0.7272 \pm 0.003$ \\
        OMIB \citep{Wu_Shao_Wang_Sun_2025}
            & $0.6752 \pm 0.008$ 
            & $0.7272 \pm 0.003$ \\
        
        \midrule 

        \multicolumn{1}{l}{\hspace{-6pt}\textit{Head-Based}} \\
        MulT \citep{Tsai_Bai_Liang_Kolter_Morency_Salakhutdinov_2019}
            & n/a 
            & n/a \\
        \midrule 

        \multicolumn{1}{l}{\hspace{-6pt}\textit{Gradient-Based}} \\
        OGM \citep{Peng_Wei_Deng_Wang_Hu_2022}
            & $0.7246 \pm 0.009$ 
            & $0.7686 \pm 0.003$ \\
        $\text{OGM}^{\text{GE}}$ \citep{Peng_Wei_Deng_Wang_Hu_2022}
            & $0.6919 \pm 0.009$ 
            & $0.7576 \pm 0.002$ \\
        DGL \citep{Wei_Luo_Luo_2025a}
            & $0.6364 \pm 0.006$ 
            & $0.7165 \pm 0.017$ \\
        ARL \citep{Wei_Luo_Luo_2025b}
            & $0.7152 \pm 0.004$ 
            & $0.7703 \pm 0.004$ \\
        MMPareto \citep{Wei_Hu_2024}
            & $0.7195 \pm 0.002$ 
            & $0.7678 \pm 0.003$ \\
        BMML \citep{Wu_Jastrzebski_Cho_Geras_2022}
            & $0.7204 \pm 0.001$ 
            & $0.7762 \pm 0.003$ \\
        G-Blend Online \citep{Wang_Tran_Feiszli_2020}
            & $0.7212 \pm 0.009$ 
            & $0.7740 \pm 0.010$ \\
        G-Blend Offline \citep{Wang_Tran_Feiszli_2020}
            & $0.7083 \pm 0.021$ 
            & $0.7625 \pm 0.010$ \\
        \midrule 

        \ac{simbamm}$^{\text{CLS}}$
            & $0.7211 \pm 0.003$ 
            & $0.7660 \pm 0.003$ \\
        \ac{simbamm}
            & n/a 
            & n/a \\

        \arrayrulecolor{black}
        
        \bottomrule
    \end{tabular}
    
\end{table}

\section{Hyperparameter Search}
\label{app:hyperparameter_search}

We use Bayesian optimization minimizing the validation loss without sweeping the batch size \citep{tuningplaybookgithub}. In the following, we list the hyperparameter search space and configurations structured by the dataset and the evaluated methods. The former refers to the encoders used, while the latter refers primarily to the remaining groups. The initial \ac{simbamm} is swept with \num{500} runs while the other methods are swept with an additional \num{100} runs.

\setminted{
  linenos,
  numbersep=4pt,
  xleftmargin=1em,
}

\subsection{Encoders Parameter Space}
The parameter spaces for the individual datasets are shown in \Cref{lst:haim,lst:symile,lst:inspect,lst:ukb,lst:mosi,lst:mosei,lst:chsims,lst:chsims2,lst:cremad}.
\begin{listing}[ht]
\begin{minted}
[
    frame=lines,
    framesep=2mm,
    linenos,
    fontsize=\scriptsize,
]
{yaml}
# Encoders
vision.vit.vit:
    values: ["vit_b_16", "vit_b_32"]
vision.vit.dropout:
    values: [0, 0.1, 0.2]
sequential.transformer.num_hidden_layers:
    values: [2, 4, 6, 8]
sequential.transformer.num_attention_heads:
    values: [2, 4, 8, 16]
sequential.transformer.intermediate_size:
    values: [256, 512, 1024, 2048]
sequential.transformer.dropout:
    values: [0, 0.1, 0.2]
\end{minted}
\caption{Hyperparameter search space configuration for MIMIC HAIM.}
\label{lst:haim}
\end{listing}

\begin{listing}[ht]
\begin{minted}
[
    frame=lines,
    framesep=2mm,
    linenos,
    fontsize=\scriptsize,
]
{yaml}
# Encoders
vision.vit.vit:
    values: ["vit_b_16", "vit_b_32"]
vision.vit.dropout:
    values: [0, 0.1, 0.2]
sequential1.transformer.num_hidden_layers:
    values: [2, 4, 6, 8]
sequential1.transformer.num_attention_heads:
    values: [2, 4, 8, 16]
sequential1.transformer.intermediate_size:
    values: [256, 512, 1024, 2048]
sequential1.transformer.dropout:
    values: [0, 0.1, 0.2]
sequential2.mlp.hidden_dims:
    values: [[256, 512, 256], [512, 1024, 512], 
            [1024, 2048, 1024]]
sequential2.mlp.hidden_dropouts:
    values: [[0, 0, 0], [0.1, 0.1, 0.1], 
            [0.2, 0.2, 0.2]]
\end{minted}
\caption{Hyperparameter search space configuration for MIMIC Symile.}
\label{lst:symile}
\end{listing}
\begin{listing}[ht]
\begin{minted}
[
    frame=lines,
    framesep=2mm,
    linenos,
    fontsize=\scriptsize,
]
{yaml}
# Encoders
vision.mlp.hidden_dims:
    values: [[1024, 512, 256], [512, 256, 128], 
            [256, 128, 64], [128, 64, 32]]
vision.mlp.hidden_dropouts:
    values: [[0.0, 0.0, 0.0], [0.1, 0.1, 0.1], 
            [0.2, 0.2, 0.2]]
ehr.mlp.hidden_dims:
    values: [[1024, 512, 256], [512, 256, 128], 
            [256, 128, 64], [128, 64, 32]]
ehr.mlp.hidden_dropouts:
    values: [[0.0, 0.0, 0.0], [0.1, 0.1, 0.1], 
            [0.2, 0.2, 0.2]]
\end{minted}
\caption{Hyperparameter search space configuration for INSPECT.}
\label{lst:inspect}
\end{listing}
\begin{listing}[ht]
\begin{minted}
[
    frame=lines,
    framesep=2mm,
    linenos,
    fontsize=\scriptsize,
]
{yaml}
# Encoders, for all 23 modalities, exemplary:
nmr.mlp.hidden_dims:
    values: [[1024, 512, 256], [512, 256, 128], 
            [256, 128, 64], [128, 64, 32]]
nmr.mlp.hidden_dropouts:
    values: [[0.0, 0.0, 0.0], [0.2, 0.2, 0.2], 
            [0.3, 0.3, 0.3], [0.4, 0.4, 0.4], 
            [0.5, 0.5, 0.5], [0.6, 0.6, 0.6], 
            [0.7, 0.7, 0.7]]
\end{minted}
\caption{Hyperparameter search space configuration for the \ac{ukb}.}
\label{lst:ukb}
\end{listing}
\begin{listing}[ht]
\begin{minted}
[
    frame=lines,
    framesep=2mm,
    linenos,
    fontsize=\scriptsize,
]
{yaml}
# Encoders
audio.transformer.num_hidden_layers:
    values: [1, 4, 8, 16]
audio.transformer.num_attention_heads:
    values: [1, 2, 4, 8]
audio.transformer.intermediate_size:
    values: [256, 512, 1024]
audio.transformer.dropout:
    values: [0.0, 0.1]
vision.transformer.num_hidden_layers:
    values: [1, 4, 8, 16]
vision.transformer.num_attention_heads:
    values: [1, 2, 4, 8]
vision.transformer.intermediate_size:
    values: [256, 512, 1024]
vision.transformer.dropout:
    values: [0.0, 0.1]
\end{minted}
\caption{Hyperparameter search space configuration for MOSI.}
\label{lst:mosi}
\end{listing}

\begin{listing}[ht]
\begin{minted}
[
    frame=lines,
    framesep=2mm,
    linenos,
    fontsize=\scriptsize,
]
{yaml}
# Encoders
audio.transformer.num_hidden_layers:
    values: [1, 4, 8, 16]
audio.transformer.num_attention_heads:
    values: [1, 2, 4, 8]
audio.transformer.intermediate_size:
    values: [256, 512, 1024]
audio.transformer.dropout:
    values: [0.0, 0.1]
vision.transformer.num_hidden_layers:
    values: [1, 4, 8, 16]
vision.transformer.num_attention_heads:
    values: [1, 2, 4, 8]
vision.transformer.intermediate_size:
    values: [256, 512, 1024]
vision.transformer.dropout:
    values: [0.0, 0.1]
\end{minted}
\caption{Hyperparameter search space configuration for MOSEI.}
\label{lst:mosei}
\end{listing}

\begin{listing}[ht]
\begin{minted}
[
    frame=lines,
    framesep=2mm,
    linenos,
    fontsize=\scriptsize,
]
{yaml}
# Encoders
audio.transformer.num_hidden_layers:
    values: [1, 4, 8, 16]
audio.transformer.num_attention_heads:
    values: [1, 2, 4, 8]
audio.transformer.intermediate_size:
    values: [256, 512, 1024]
vision.transformer.num_hidden_layers:
    values: [1, 4, 8, 16]
vision.transformer.num_attention_heads:
    values: [1, 2, 4, 8]
vision.transformer.intermediate_size:
    values: [256, 512, 1024]
\end{minted}
\caption{Hyperparameter search space configuration for CHSIMS.}
\label{lst:chsims}
\end{listing}
\begin{listing}[ht]
\begin{minted}
[
    frame=lines,
    framesep=2mm,
    linenos,
    fontsize=\scriptsize,
]
{yaml}
# Encoders
audio.transformer.num_hidden_layers:
    values: [1, 4, 8, 16]
audio.transformer.num_attention_heads:
    values: [1, 2, 4, 8]
audio.transformer.intermediate_size:
    values: [256, 512, 1024]
vision.transformer.num_hidden_layers:
    values: [1, 4, 8, 16]
vision.transformer.num_attention_heads:
    values: [1, 2, 4, 8]
vision.transformer.intermediate_size:
    values: [256, 512, 1024]
\end{minted}
\caption{Hyperparameter search space configuration for CHSIMS2.}
\label{lst:chsims2}
\end{listing}
\begin{listing}[ht]
\begin{minted}
[
    frame=lines,
    framesep=2mm,
    linenos,
    fontsize=\scriptsize,
]
{yaml}
# Encoders
# Standard PyTorch ResNet18s
\end{minted}
\caption{Hyperparameter search space configuration for CREMA-D.}
\label{lst:cremad}
\end{listing}

\subsection{Methods Parameter Space}
The parameter spaces for the individual methods are shown in \Cref{lst:simbamm,lst:imder,lst:mmp,lst:shaspec,lst:regbn,lst:aug,lst:mbt,lst:lmf,lst:pdf,lst:coupled_mamba,lst:omib,lst:mult,lst:ogm,lst:dgl,lst:arl,lst:mmpareto,lst:bmml,lst:gblend}.
\begin{listing}[ht]
\begin{minted}
[
    frame=lines,
    framesep=2mm,
    linenos,
    fontsize=\scriptsize,
]
{yaml}
# Head 
head_transformer.d_model:
    values: [32, 64, 128, 256, 512]
head_transformer.dim_feedforward:
    values: [256, 512, 1024, 2048]
head_transformer.dropout:
    values: [0, 0.1, 0.2]
head_transformer.nhead:
    values: [4, 8, 16]
head_transformer.num_layers:
    values: [2, 4, 6, 8]
# Optimizer
optimizer.lr:
    min: 0.000001
    max: 0.1
    distribution: "log_uniform_values"
optimizer.warmup_steps:
    values: [0, 100, 200, 500, 1000]
optimizer.weight_decay:
    values: [0, 0.1, 0.01, 0.001]
\end{minted}
\caption{Hyperparameter search space configuration for \ac{simbamm}.}
\label{lst:simbamm}
\end{listing}
\begin{listing}[ht]
\begin{minted}
[
    frame=lines,
    framesep=2mm,
    linenos,
    fontsize=\scriptsize,
]
{yaml}
# Optimizer
optimizer.lr:
    min: 0.000001
    max: 0.1
    distribution: "log_uniform_values"
optimizer.warmup_steps:
    values: [0, 100, 200, 500, 1000]
optimizer.weight_decay:
    values: [0, 0.1, 0.01, 0.001]
# IMDer
imder.beta:
    min: 0.01
    max: 1.0
    distribution: "log_uniform_values"
ddpms.n_steps:
    values: [10, 20, 30, 40, 50, 100]
ddpms.d_model: # fixed w.r.t. Transformer d_model
    value: 256
ddpms.hidden_dim:
    values: [32, 64, 128, 256, 512, 1024]
ddpms.num_layers:
    values: [1, 2, 3, 4, 5, 6, 7, 8]
ddpms.dropout:
    values: [0.0, 0.1, 0.2]
ddpms.nhead:
    values: [1, 2, 4, 8, 16, 32]
\end{minted}
\caption{Hyperparameter search space configuration for IMDer.}
\label{lst:imder}
\end{listing}
\begin{listing}[ht]
\begin{minted}
[
    frame=lines,
    framesep=2mm,
    linenos,
    fontsize=\scriptsize,
]
{yaml}
# Optimizer
optimizer.lr:
    min: 0.000001
    max: 0.1
    distribution: "log_uniform_values"
optimizer.warmup_steps:
    values: [0, 100, 200, 500, 1000]
optimizer.weight_decay:
    values: [0, 0.1, 0.01, 0.001]
# MMP
mmp.proj_mlp.hidden_dim:
    values: [128, 256, 512, 1024]
mmp.proj_mlp.dropout:
    values: [0.0, 0.1, 0.2]
mmp.attn_steps.dropout:
    values: [0.0, 0.1, 0.2]
mmp.attn_steps.nhead:
    values: [1, 2, 4, 8, 16]
mmp.num_aggregated_tokens:
    values: [2, 4, 8, 16, 32]
mmp.loss_alignment_alpha:
    min: 0.001
    max: 1.0
    distribution: "log_uniform_values"
\end{minted}
\caption{Hyperparameter search space configuration for MMP.}
\label{lst:mmp}
\end{listing}
\begin{listing}[ht]
\begin{minted}
[
    frame=lines,
    framesep=2mm,
    linenos,
    fontsize=\scriptsize,
]
{yaml}
# Optimizer
optimizer.lr:
    min: 0.000001
    max: 0.1
    distribution: "log_uniform_values"
optimizer.warmup_steps:
    values: [0, 100, 200, 500, 1000]
optimizer.weight_decay:
    values: [0, 0.1, 0.01, 0.001]
# ShaSpec
shaspec.loss_alpha:
    min: 0.01
    max: 1.0
    distribution: "log_uniform_values"
shaspec.loss_beta:
    min: 0.001
    max: 0.1
    distribution: "log_uniform_values"
\end{minted}
\caption{Hyperparameter search space configuration for ShaSpec.}
\label{lst:shaspec}
\end{listing}
\begin{listing}[ht]
\begin{minted}
[
    frame=lines,
    framesep=2mm,
    linenos,
    fontsize=\scriptsize,
]
{yaml}
# Optimizer
optimizer.lr:
    min: 0.000001
    max: 0.1
    distribution: "log_uniform_values"
optimizer.warmup_steps:
    values: [0, 100, 200, 500, 1000]
optimizer.weight_decay:
    values: [0, 0.1, 0.01, 0.001]
# RegBN
rbn.reference_modality_idx:  # 0-(M-1)
    values: [0, 1, 2]
rbn.momentum:
    values: [0.1, 0.01, 0.05]
rbn.sigma_THR:
    values: [0.1, 0.3, 0.5]
rbn.sigma_MIN:
    values: [0.01, 0.1, 0.2]
rbn.affine:
    values: [True, False]
\end{minted}
\caption{Hyperparameter search space configuration for RegBN.}
\label{lst:regbn}
\end{listing}
\begin{listing}[ht]
\begin{minted}
[
    frame=lines,
    framesep=2mm,
    linenos,
    fontsize=\scriptsize,
]
{yaml}
# Optimizer
optimizer.lr:
    min: 0.000001
    max: 0.1
    distribution: "log_uniform_values"
optimizer.warmup_steps:
    values: [0, 100, 200, 500, 1000]
optimizer.weight_decay:
    values: [0, 0.1, 0.01, 0.001]
# AUG
aug.merge_alphas:
    values: [[0.33,0.33,0.33]]  # 1/M
aug.check_interval:
    values: [1, 5, 10, 20]
aug.threshold:
    min: 0.01
    max: 1.0
    distribution: "log_uniform_values"
aug.confidence_coeff:
    min: 0.01
    max: 10.0
    distribution: "log_uniform_values"
aug.lambda_smooth:
    min: 0.01
    max: 1.0
    distribution: "log_uniform_values"
\end{minted}
\caption{Hyperparameter search space configuration for AUG.}
\label{lst:aug}
\end{listing}
\begin{listing}[ht]
\begin{minted}
[
    frame=lines,
    framesep=2mm,
    linenos,
    fontsize=\scriptsize,
]
{yaml}
# Optimizer
optimizer.lr:
    min: 0.000001
    max: 0.1
    distribution: "log_uniform_values"
optimizer.warmup_steps:
    values: [0, 100, 200, 500, 1000]
optimizer.weight_decay:
    values: [0, 0.1, 0.01, 0.001]
# MBT
bottleneck.num_bottlenecks:
    values: [1, 2, 4, 8, 16, 32]
bottleneck.layers:
    values: [1, 2, 4, 8, 16, 32]
bottleneck.dim_feedforward:
    values: [128, 256, 512, 1024, 2048]
bottleneck.nhead:
    values: [1, 2, 4, 8, 16, 32]
bottleneck.dropout:
    values: [0.0, 0.1, 0.2, 0.3, 0.4]
\end{minted}
\caption{Hyperparameter search space configuration for MBT.}
\label{lst:mbt}
\end{listing}
\begin{listing}[ht]
\begin{minted}
[
    frame=lines,
    framesep=2mm,
    linenos,
    fontsize=\scriptsize,
]
{yaml}
# Optimizer
optimizer.lr:
    min: 0.000001
    max: 0.1
    distribution: "log_uniform_values"
optimizer.warmup_steps:
    values: [0, 100, 200, 500, 1000]
optimizer.weight_decay:
    values: [0, 0.1, 0.01, 0.001]
# LMF
low_rank_matrix_fusion.rank:
    values: [2, 10, 20, 50, 100, 200, 250]
\end{minted}
\caption{Hyperparameter search space configuration for LMF.}
\label{lst:lmf}
\end{listing}
\begin{listing}[ht]
\begin{minted}
[
    frame=lines,
    framesep=2mm,
    linenos,
    fontsize=\scriptsize,
]
{yaml}
# Optimizer
optimizer.lr:
    min: 0.000001
    max: 0.1
    distribution: "log_uniform_values"
optimizer.warmup_steps:
    values: [0, 100, 200, 500, 1000]
optimizer.weight_decay:
    values: [0, 0.1, 0.01, 0.001]
# PDF
pdf.loss_weight:
    min: 0.01
    max: 1.0
    distribution: "log_uniform_values"
pdf.unimodal_loss_weight:
    min: 0.01
    max: 1.0
    distribution: "log_uniform_values"
pdf.p_head.hidden_dims:
    values: [[128], [256], [512], [1024], 
            [128, 256], [128, 256, 512], 
            [128, 256, 512, 1024]]
pdf.p_head.dropout:
    values: [0.0, 0.1, 0.2]
\end{minted}
\caption{Hyperparameter search space configuration for PDF.}
\label{lst:pdf}
\end{listing}
\begin{listing}[ht]
\begin{minted}
[
    frame=lines,
    framesep=2mm,
    linenos,
    fontsize=\scriptsize,
]
{yaml}
# Optimizer
optimizer.lr:
    min: 0.000001
    max: 0.1
    distribution: "log_uniform_values"
optimizer.warmup_steps:
    values: [0, 100, 200, 500, 1000]
optimizer.weight_decay:
    values: [0, 0.1, 0.01, 0.001]
# Coupled Mamba
coupled_ssm.d_model:
    values: [32, 64, 128, 256]
coupled_ssm.num_layers:
    values: [2, 3, 4, 5, 6, 7, 8]
coupled_ssm.d_state:
    values: [8, 16, 32, 64, 128]
coupled_ssm.d_conv:
    values: [2, 4, 8, 16]
coupled_ssm.expand:
    values: [1, 2, 4, 8]
\end{minted}
\caption{Hyperparameter search space configuration for Coupled Mamba.}
\label{lst:coupled_mamba}
\end{listing}
\begin{listing}[ht]
\begin{minted}
[
    frame=lines,
    framesep=2mm,
    linenos,
    fontsize=\scriptsize,
]
{yaml}
# Optimizer
optimizer.lr:
    min: 0.000001
    max: 0.1
    distribution: "log_uniform_values"
optimizer.warmup_steps:
    values: [0, 100, 200, 500, 1000]
optimizer.weight_decay:
    values: [0, 0.1, 0.01, 0.001]
# OMIB
omib.beta:
    min: 0.00001
    max: 0.1
    distribution: "log_uniform_values"
omib.cross_attn_network.num_layers:
    values: [1, 2, 4, 8, 16]
omib.cross_attn_network.dropout:
    values: [0.0, 0.1, 0.2]
omib.cross_attn_network.num_heads:
    values: [1, 2, 4, 8, 16]
omib.cross_attn_network.dim_feedforward:
    values: [128, 256, 512, 1024]
omib.warmup_epochs:
    values: [0, 1, 2, 4, 8]
\end{minted}
\caption{Hyperparameter search space configuration for OMIB.}
\label{lst:omib}
\end{listing}
\begin{listing}[ht]
\begin{minted}
[
    frame=lines,
    framesep=2mm,
    linenos,
    fontsize=\scriptsize,
]
{yaml}
# Optimizer
optimizer.lr:
    min: 0.000001
    max: 0.1
    distribution: "log_uniform_values"
optimizer.warmup_steps:
    values: [0, 100, 200, 500, 1000]
optimizer.weight_decay:
    values: [0, 0.1, 0.01, 0.001]
# MulT
crossmodal_transformer.d_model:
    values: [256]  # = head_transformer.d_model
crossmodal_transformer.num_layers:
    values: [1, 2, 4, 8, 16]
crossmodal_transformer.dim_feedforward:
    values: [128, 256, 512, 1024]
crossmodal_transformer.nhead:
    values: [1, 2, 4, 8, 16]
crossmodal_transformer.dropout:
    values: [0.0, 0.1, 0.2, 0.5]
\end{minted}
\caption{Hyperparameter search space configuration for MulT.}
\label{lst:mult}
\end{listing}
\begin{listing}[ht]
\begin{minted}
[
    frame=lines,
    framesep=2mm,
    linenos,
    fontsize=\scriptsize,
]
{yaml}
# Optimizer
optimizer.lr:
    min: 0.000001
    max: 0.1
    distribution: "log_uniform_values"
optimizer.warmup_steps:
    values: [0, 100, 200, 500, 1000]
optimizer.weight_decay:
    values: [0, 0.1, 0.01, 0.001]
# OGM
ogm.alpha:
    min: 0.01
    max: 1.0
    distribution: "log_uniform_values"
ogm.use_ge:
    values: [False]  # True for OGM-GE
\end{minted}
\caption{Hyperparameter search space configuration for OGM and OGM-GE.}
\label{lst:ogm}
\end{listing}
\begin{listing}[ht]
\begin{minted}
[
    frame=lines,
    framesep=2mm,
    linenos,
    fontsize=\scriptsize,
]
{yaml}
# Optimizer
optimizer.lr:
    min: 0.000001
    max: 0.1
    distribution: "log_uniform_values"
optimizer.warmup_steps:
    values: [0, 100, 200, 500, 1000]
optimizer.weight_decay:
    values: [0, 0.1, 0.01, 0.001]
# DGL
dgl.unimodal_loss_weight:
    min: 0.001
    max: 1.0
    distribution: "log_uniform_values"
\end{minted}
\caption{Hyperparameter search space configuration for DGL.}
\label{lst:dgl}
\end{listing}
\begin{listing}[ht]
\begin{minted}
[
    frame=lines,
    framesep=2mm,
    linenos,
    fontsize=\scriptsize,
]
{yaml}
# Optimizer
optimizer.lr:
    min: 0.000001
    max: 0.1
    distribution: "log_uniform_values"
optimizer.warmup_steps:
    values: [0, 100, 200, 500, 1000]
optimizer.weight_decay:
    values: [0, 0.1, 0.01, 0.001]
# ARL
arl.unimodal_loss_weight:
    min: 0.01
    max: 1.0
    distribution: "log_uniform_values"
arl.temperature:
    min: 0.01
    max: 1.0
    distribution: "log_uniform_values"
\end{minted}
\caption{Hyperparameter search space configuration for ARL.}
\label{lst:arl}
\end{listing}
\begin{listing}[ht]
\begin{minted}
[
    frame=lines,
    framesep=2mm,
    linenos,
    fontsize=\scriptsize,
]
{yaml}
# Optimizer
optimizer.lr:
    min: 0.000001
    max: 0.1
    distribution: "log_uniform_values"
optimizer.warmup_steps:
    values: [0, 100, 200, 500, 1000]
optimizer.weight_decay:
    values: [0, 0.1, 0.01, 0.001]
# MMPareto
mmpareto.unimodal_loss_weight:
    min: 0.01
    max: 1.0
    distribution: "log_uniform_values"
mmpareto.gamma:
    min: 0.01
    max: 1.0
    distribution: "log_uniform_values"
\end{minted}
\caption{Hyperparameter search space configuration for MMPareto.}
\label{lst:mmpareto}
\end{listing}
\begin{listing}[ht]
\begin{minted}
[
    frame=lines,
    framesep=2mm,
    linenos,
    fontsize=\scriptsize,
]
{yaml}
# Optimizer
optimizer.lr:
    min: 0.000001
    max: 0.1
    distribution: "log_uniform_values"
optimizer.warmup_steps:
    values: [0, 100, 200, 500, 1000]
optimizer.weight_decay:
    values: [0, 0.1, 0.01, 0.001]
# BMML
bmml.unimodal_loss_weight:
    min: 0.01
    max: 1.0
    distribution: "log_uniform_values"
bmml.alpha:
    min: 0.001
    max: 0.1
    distribution: "log_uniform_values"
bmml.q:
    values: [3, 5, 10]
bmml.bmml_momentum:
    values: [0.9, 0.99, 0.999]
bmml.warmup_epochs:
    values: [0, 1, 2, 5, 10]
\end{minted}
\caption{Hyperparameter search space configuration for BMML.}
\label{lst:bmml}
\end{listing}
\begin{listing}[ht]
\begin{minted}
[
    frame=lines,
    framesep=2mm,
    linenos,
    fontsize=\scriptsize,
]
{yaml}
# Optimizer
optimizer.lr:
    min: 0.000001
    max: 0.1
    distribution: "log_uniform_values"
optimizer.warmup_steps:
    values: [0, 100, 200, 500, 1000]
optimizer.weight_decay:
    values: [0, 0.1, 0.01, 0.001]
# GBlend
gblend.mode:
    values: ['offline']  # Online 
gblend.update_freq:
    values: [1, 5, 10]
gblend.lookahead_epochs:
    values: [1, 3, 5]
\end{minted}
\caption{Hyperparameter search space configuration for G-Blend Online/Offline.}
\label{lst:gblend}
\end{listing}

%\clearpage
%\newpage
%\input{checklist.tex}

\end{document}